\documentclass[preprint,review,12pt]{elsarticle}

\usepackage{url}
\usepackage[utf8]{inputenc} 
\usepackage[T1]{fontenc}    
\usepackage{url}            
\usepackage{booktabs}       
\usepackage{nicefrac}       
\usepackage{microtype}      
\usepackage{subcaption}
\usepackage{algorithm}
\usepackage{algpseudocode}
\usepackage{times}
\usepackage{epsfig}
\usepackage{graphicx}
\usepackage{amssymb}

\usepackage{multirow}
\usepackage{enumitem}
\usepackage{comment}
\usepackage{booktabs}

\usepackage{amsmath}
\usepackage{amssymb}
\usepackage{booktabs}
\usepackage{multirow}
\usepackage{xcolor}
\usepackage{duckuments}
\usepackage{float}
\usepackage{mathtools}
\usepackage{amssymb}
\usepackage{soul}
\usepackage{comment}
\usepackage{makecell}

\usepackage{hyperref}
\usepackage{color}
\usepackage{tabularx}
\usepackage{booktabs}
\usepackage{multirow, array}
\usepackage{tabularx, colortbl}
\usepackage{multicol, blindtext}
\usepackage{interval}
\usepackage{bbold}
\usepackage{wrapfig}

\usepackage{amsmath,amsfonts,bm}

\def\eqref#1{equation~\ref{#1}}
\def\Eqref#1{Equation~\ref{#1}}

\def\vs{{\bm{s}}}

\def\vu{{\bm{u}}}

\def\mE{{\bm{E}}}

\def\mW{{\bm{W}}}

\DeclareMathAlphabet{\mathsfit}{\encodingdefault}{\sfdefault}{m}{sl}
\SetMathAlphabet{\mathsfit}{bold}{\encodingdefault}{\sfdefault}{bx}{n}

\newcommand{\Ls}{\mathcal{L}}

\newcommand{\softmax}{\mathrm{softmax}}

\DeclareMathOperator{\squash}{squash}

\journal{Pattern Recognition}

\definecolor{ForestGreen}{RGB}{0,0,0}
\definecolor{ForestGreen2}{RGB}{0,0,0}
\definecolor{LightGreen}{HTML}{99ff99}
\definecolor{Yellow}{HTML}{FFF8AD}

\begin{document}



\begin{frontmatter}



\title{Capsule Networks Do Not Need to Model Everything}


\author[inst1]{Riccardo\corref{cor1} Renzulli}
\cortext[cor1]{Corresponding author.}
\ead{riccardo.renzulli@unito.it}

\affiliation[inst1]{organization={University of Turin, Computer Science Department},
            country={Italy}}

\author[inst2]{Enzo Tartaglione}
\author[inst1]{Marco Grangetto}

\affiliation[inst2]{organization={LTCI, Télécom Paris, Institut Polytechnique de Paris},
            country={France}}

\begin{abstract}
Capsule networks are biologically inspired neural networks that group neurons into vectors called capsules, each explicitly representing an object or one of its parts. The routing mechanism connects capsules in consecutive layers, forming a hierarchical structure between parts and objects, also known as a parse tree. 
\textcolor{ForestGreen}{Capsule networks often attempt to model all elements in an image, requiring large network sizes to handle complexities such as intricate backgrounds or irrelevant objects. However, this comprehensive modeling leads to increased parameter counts and computational inefficiencies. 
Our goal is to enable capsule networks to focus only on the object of interest, reducing the number of parse trees.} We accomplish this with REM (Routing Entropy Minimization), a technique that minimizes the entropy of the parse tree-like structure. REM drives the model parameters distribution towards low entropy configurations through a pruning mechanism, significantly reducing the generation of intra-class parse trees. This empowers capsules to learn more stable and succinct representations with fewer parameters and negligible performance loss.
\end{abstract}



\begin{keyword}
deep learning \sep capsule networks \sep routing \sep parse trees
\end{keyword}

\end{frontmatter}


\section{Introduction}
\label{sec:intro}
Capsule Networks (CapsNets)~\citep{hinton-dynamic} were introduced to overcome the shortcomings of Convolutional Neural Networks (CNNs). In fact,
CNNs lose the spatial relationships between objects and their parts in the input image because of max pooling layers, which progressively drop spatial information~\citep{hinton-dynamic}. However, modeling such spatial relationships can improve the robustness of the network to unseen viewpoints without relying on massive data augmentations~\citep{hinton-dynamic}. CapsNets encode objects into groups of neurons called capsules, and they can learn hierarchical structures from data thanks to a routing algorithm. This mechanism provides a way to decompose a scene into objects and their parts, representing their spatial relationships into a hierarchical structure called \textit{parse tree}. Due to this strong inductive bias, CapsNets are more robust to viewpoint changes and affine transformations than traditional CNNs.
 
Despite their potential advantages, CapsNets have not yet seen widespread adoption in the industry or research community compared to traditional CNNs or recently introduced Vision Transformers (ViTs)~\cite{dosovitskiy2021an, liu2021Swin}. The lack of standardized architectures and pre-trained models might be a contributing factor. However, continued research and advances in CapsNets are essential to fully harness their potential and address their limitations. 

Our paper \textcolor{ForestGreen}{is motivated} by a drawback of CapsNets identified by \citet{hinton-dynamic}: these networks perform better when they can model all elements present in an image; however, larger network sizes are required to handle image clutter, such as intricate backgrounds or complex objects. Therefore, the entropy of the connections of the parse trees is high.  Recently, \citet{global_routing} showed that a colored background in the input image can generate invalid background voting capsules that can reduce CapsNet performance. However, their proposed solution includes the addition of additional trainable modules to filter out irrelevant capsules. Our goal is not to achieve state-of-the-art performance in computer vision tasks with CapsNets, but rather to extract fewer and more succinct part-whole relationships from existing capsule models. In the literature, many pruning methods were applied to CNNs to reduce the complexity of networks, enforcing sparse topologies~\citep{enzo-neurips, sparsevd, louizos}. These considerations lead us to the following question: \textit{is it possible to aid the extraction of more discriminative features (namely fewer parse trees), tailoring pruning and quantization approaches?}

This work introduces REM (Routing Entropy Minimization), which moves some steps toward extracting more succinct relationships from CapsNets with negligible performance loss and fewer parameters. To encourage this, we impose sparsity and entropy constraints. In low pruning regimes, noisy couplings cause the entropy to increase considerably. In contrast, in high-sparsity regimes, pruning can effectively reduce the overall entropy of the connections of the parse tree-like structure encoded in a CapsNet. 
We collect the coupling coefficients, studying their frequency and cardinality, observing lower intra-class conditional entropy: the pruned version \emph{adds a missing explicit prior} in the routing mechanism, grounding the coupling of the unused capsules and disallowing fluctuations under the same baseline performance on the validation/test set. This implies that the parse trees are significantly less, hence more stable for the pruned models, focusing on more relevant features (per class) selected from the routing mechanism. We also introduce two visualization methods of the parse trees based on the last type of capsule layer.

Our main contributions can be summarized as follows:
\begin{itemize}
    \item we provide novel insights on CapsNets and their ability to extract part-whole relationships, also known as parse trees, showing that existing models typically focus also on irrelevant entities or noisy backgrounds (Section ~\ref{sec:vis-rem}); 
    \item we design a novel technique called REM (Section ~\ref{sec:rem}) that exploits pruning (Section ~\ref{sec:pruning}) and quantization (Section ~\ref{sec:quantizatino}) approaches to extract more succinct representations in CapsNets. We define a quantitative metric based on the entropy to define the part-whole relationships (Section ~\ref{sec:entropy}). Our extensive experiments (Section ~\ref{sec:exp}) show that REM can be plugged into any capsule architecture to lower the entropy on many datasets successfully;
    \item we propose two novel visualization methods of the extracted parse trees, based on how capsules in the last layer are organized, to delve into the hierarchical structures extracted by CapsNets (Section~\ref{sec:extract_salmap}). With REM, we show that CapsNets with few trainable parameters do not need to model irrelevant objects or noise in the images (Section ~\ref{sec:vis-rem}) to achieve high generalization. 
\end{itemize}

\noindent The rest of the paper is organized as follows: In Section~\ref{sec:sota}, we introduce the basic concepts of CapsNets and related work; in Section~\ref{sec:method} we describe our REM technique and the two novel visualization methods; in Section~\ref{sec:results} we show the effectiveness of our approach on many datasets and architectures; and finally we discuss the conclusion of our work. Appendices are provided in the supplementary material. 
\section{Background and related work}\label{sec:sota}
This section first describes the fundamental aspects of CapsNets, focusing on the first routing algorithm introduced by~\citet{hinton-dynamic}, commonly known as dynamic routing. Then, we review the literature, especially related to sparsity in CapsNets.

\subsection{Capsule networks fundamentals}\label{sec:capsnets_fund}
\begin{figure}[ht]
\centering
\includegraphics[width=1\textwidth]{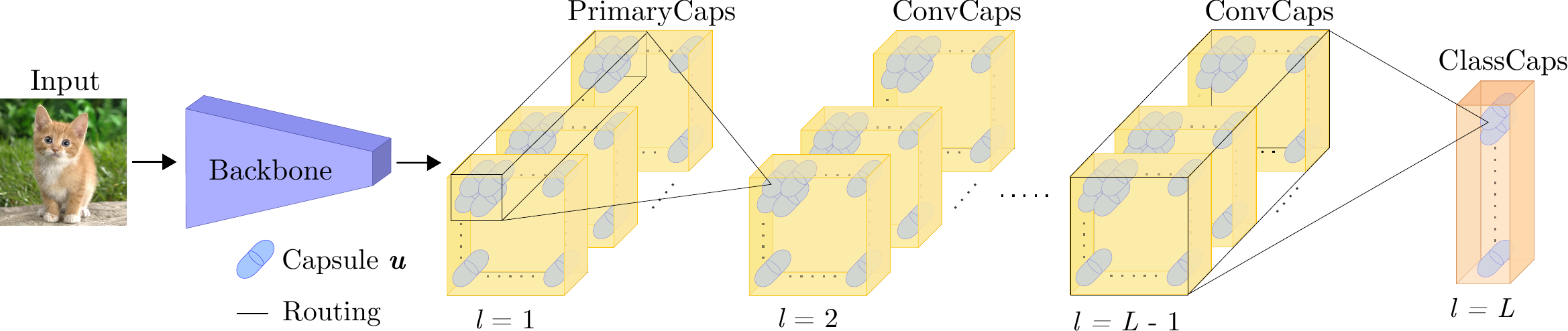}			
\caption{An example of a general CapsNet architecture composed by a convolutional backbone, a PrimaryCaps layer (which is a convolutional layer with squash activation and then reshaping), several ConvCaps layers (where a capsule in layer $l$ is computed using only a subset of capsules in layer $l-1$), and a ClassCaps (or FcCaps, where each capsule in layer $l$ is computed using all the capsules in layer $l-1$) layer.\label{fig:capsnet_encoder}}
\end{figure}

 CapsNets group neurons into \textit{capsules}, namely activity vectors $\vu$, where each capsule accounts for an object or one of its parts. Each element of these vectors accounts for different properties of the object, such as its pose and other features like color, deformation, etc. In this work, we also use the term pose to refer to the activity vector. The magnitude $\lVert \vu \rVert_2$ of a capsule stands for the probability of the existence of that object in the image. 
 A CapsNet consists of several capsule layers $l$ stacked on top of a convolutional backbone, where $l \in \{1, 2,..., L\}$. The set of capsules in layer $l$ is denoted as $\Omega^{[l]}$. There are three types of capsule layers: primary capsules (PrimaryCaps, the first one, built upon convolutional layers), convolutional capsules (ConvCaps, each capsule has its own receptive field), and fully-connected capsules (FcCaps, each capsule is connected with all the capsules in the previous layer).
 Typically, a CapsNet comprises at least two capsule layers, PrimaryCaps and FcCaps (also called ClassCaps when it is the last layer, with one output capsule for each object class). Figure~\ref{fig:capsnet_encoder} shows an example of a CapsNet architecture.

\textit{Assembling} capsules in layer $l$ to compute capsules in layer $l+1$ is a two-stage process: part-whole transformation and hierarchical routing.
During the first phase, each capsule $i$ in layer $l$, whose pose is denoted as $\vu^{[l]}_i$, makes a prediction $\hat{\vu}^{[l+1]}_{i,j}$, thanks to a transformation matrix $\mW^{[l]}_{i,j}$, for the pose $\vu^{[l+1]}_j$ of an upper layer capsule $j$
 \begin{equation}
     \hat{\vu}^{[l+1]}_{i,j} = \mW^{[l]}_{i,j}\vu^{[l]}_i.
 \end{equation} 
Then, during the second phase, the unnormalized pose $\vs^{[l+1]}_j$ is computed as the weighted average of votes $\hat{\vu}^{[l+1]}_{i,j}$
\begin{equation}\label{eq:sj}
    \vs^{[l+1]}_j = \sum\limits_{i} c^{[l]}_{i,j} \hat{\vu}^{[l+1]}_{i,j},
\end{equation}
where $c^{[l]}_{i,j}$ are the coupling coefficients between a capsule $i$ in layer $l$ and a capsule $j$ in layer $l+1$.
The pose $\vu^{[l+1]}_j$ is then defined as the normalized ``squashed'' $\vs^{[l+1]}_j$
 \begin{equation}\label{eq:squashing}
	\vu^{[l+1]}_j = \squash(\vs^{[l+1]}_j) = \frac{\|\vs^{[l+1]}_j\|^{2}}{1 + \|\vs^{[l+1]}_j\|^{2}} \frac{\vs^{[l+1]}_j}{\|\vs^{[l+1]}_j\|},
\end{equation}
whose magnitude lies in the range [0, 1). 
The coupling coefficients are computed dynamically and depend on the input.
They are determined by a ``routing softmax'' activation function, whose initial logits $b^{[l]}_{i,j}$ are the log prior probabilities that the $i$-th capsule should be coupled to the $j$-th one 
\begin{equation}\label{eq:softmax}
		c^{[l]}_{i,j} = \softmax(b^{[l]}_{i,j}) = \frac{\exp\left(b^{[l]}_{i,j}\right)}{\sum\limits_{k} \exp\left(b^{[l]}_{i,k}\right)}. \\
\end{equation}
At the first step of the routing algorithm, they are equal, and then they are refined by measuring the agreement (defined as the scalar product) between the pose $\vu^{[l+1]}_j$ and the prediction $\hat{\vu}^{[l+1]}_{i,j}$ for a given input. 
At each iteration, the update rule for the logits is
\begin{equation}\label{eq:update-rule}
    b^{[l]}_{i,j} \gets b^{[l]}_{i,j} + \vu^{[l+1]}_j\hat{\vu}^{[l+1]}_{i,j}.
\end{equation}
The steps defined in \Eqref{eq:sj}, \Eqref{eq:squashing}, \Eqref{eq:softmax}, \Eqref{eq:update-rule} are repeated for the $r$ iterations of the routing algorithm. \textcolor{ForestGreen2}{Algorithm~\ref{alg:capsule_routing} shows the pseucode of the routing mechanism.}
\begin{algorithm}[ht]
\caption{\textcolor{ForestGreen2}{Capsule Routing Algorithm}}
\label{alg:capsule_routing}
\begin{algorithmic}[1]
\Require \textcolor{ForestGreen2}{Votes $\hat{\vu}^{[l+1]}_{i,j}$, initial logits $b^{[l]}_{i,j}$, routing iterations $r$}
\Ensure \textcolor{ForestGreen2}{Capsules in layer $(l+1)$: $\vu^{[l+1]}_j \in \Omega^{[l+1]}$}
\For{\textcolor{ForestGreen2}{$r$ routing iterations}}
    \State \textcolor{ForestGreen2}{$c^{[l]}_{i,j} = \softmax(b^{[l]}_{i,j})$} \Comment{\textcolor{ForestGreen2}{Compute couplings coefficients}}
    \State \textcolor{ForestGreen2}{$\vs^{[l+1]}_j = \sum\limits_{i} c^{[l]}_{i,j} \hat{\vu}^{[l+1]}_{i,j}$} 
    \Comment{\textcolor{ForestGreen2}{Compute weighted votes}} 
    \State \textcolor{ForestGreen2}{$\vu^{[l+1]}_j = \squash(\vs^{[l+1]}_j)$} \Comment{\textcolor{ForestGreen2}{Apply squashing}}
    \State \textcolor{ForestGreen2}{$b^{[l]}_{i,j} \gets b^{[l]}_{i,j} + \vu^{[l+1]}_j \cdot \hat{\vu}^{[l+1]}_{i,j}$} \Comment{\textcolor{ForestGreen2}{Update logits based on agreement}}
\EndFor
\State \textcolor{ForestGreen2}{\Return $\vu^{[l+1]}_j$ for all capsules in $\Omega^{[l+1]}$}
\end{algorithmic}
\end{algorithm}

\begin{figure}
    \centering
    \includegraphics[width=1\columnwidth]{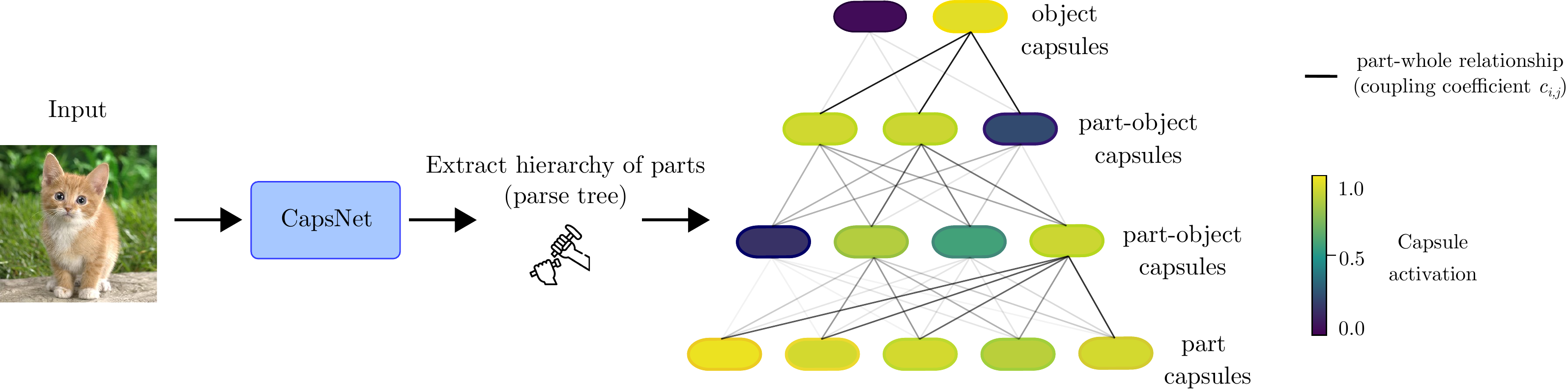}		
    \caption{Toy example of the hierarchical structure, called parse tree, extracted by a CapsNet. \textcolor{ForestGreen}{Each parse tree organize capsules in parts and objects.}\label{fig:parse_tree}}
\end{figure}

As mentioned in Section~\ref{sec:intro}, the goal of the routing algorithm is to decompose the scene into objects and parts. We say that a hierarchy of parts is \textit{carved out} of a CapsNet like a sculpture is carved from a rock. This hierarchical structure is also known as \emph{parse tree}, where each node in a layer $l$ is a capsule that encodes an entity of the input image, such as an object or one of its parts, and each connection is the coupling coefficient $c_{i,j}^{[l]}$ between capsules $i$ and $j$, which represent the part-whole or child-parent relationships between the two nodes. Figure~\ref{fig:parse_tree} depicts an example of the parse tree.
\citet{hinton-dynamic} replaced the cross-entropy loss with a \textit{margin loss}: the key idea is that the output capsule for the predicted class should have a long instantiation vector if and only if an object of that class is present in the input image.

In the following sections, for simplicity of notation, we assume there are only two capsule layers unless otherwise specified. Therefore, we suppress the layer index, and we refer to the number of capsules of the last two layers as $|\Omega^{[L-1]}| = I$ and $|\Omega^{[L]}| = J$, respectively. 

\subsection{Capsule networks follow-ups} 
Capsule networks were first introduced by ~\citet{hinton-dynamic}, and since then, much work has been done to improve the routing stage and build deeper models. Regarding the routing algorithm, \citet{hinton-em} replaces the dynamic routing with Expectation-Maximization, adopting matrix capsules instead of vector capsules. \citet{Wang2018AnOV} model the routing strategy as an optimization problem. \citet{nn-encaps} use master and aide branches to reduce the complexity of the routing process. \citet{peer2019gammacapsules} use inverse distances instead of the dot product to compute the agreements between capsules to increase their transparency and robustness against adversarial attacks. 
\citet{ribeiro2020capsule} propose a new routing algorithm derived from Variational Bayes for fitting a mixture of transforming Gaussians.
\citet{subspacecaps} model entities through a group of capsule subspaces without any form of routing. \textcolor{ForestGreen}{\citet{mandal} propose a novel two-phase dynamic routing protocol that computes agreements between neurons at various layers for micro and macro-level features, following a hierarchical learning paradigm.}
Since the CapsNet model introduced by \citet{hinton-dynamic} is a shallow network, several works attempted to build deep CapsNets. 
 \citet{deepcaps} propose a deep capsule network architecture that uses a novel 3D convolution-based dynamic routing algorithm to improve the performance of CapsNets for more complex image datasets. \citet{res-caps} introduce residual connections to train deeper capsule networks. \textcolor{ForestGreen2}{\citet{rescaps} further introduce residual pose routing to reduce the routing computation complexity and avoid gradient vanishing due to its residual learning framework.} 

\subsection{Sparse capsule networks} 
A naive solution to reduce uncertainty within the routing algorithm is to run more iterations. As shown by~\citet{paik} and \citet{gu}, the routing algorithms tend to overly polarize the link strengths, namely a simple route in which each input capsule sends its output to only one capsule and all other routes are suppressed. On the one hand, this behavior is desirable because the routing algorithm computes binary decisions to either connect or disconnect objects and parts. On the other hand, running many iterations is computationally expensive, and it is only helpful in the case of networks with few parameters, as demonstrated by \citet{renzulli}. \citet{sparse-caps} trained CapsNets in an unsupervised setting, showing that the routing algorithm no longer discriminates among capsules: the coupling coefficients collapse to the same value. Therefore, they sparsify latent capsule layer activities by masking output capsules according to a custom ranking function. 
\citet{hinton-stacked-caps} impose sparsity and entropy constraints into capsules, but they do not employ an iterative routing mechanism. \citet{ladder-caps} introduced a structured pruning layer called ladder capsule layers, which removes irrelevant capsules, namely capsules with low activities. \citet{3D-Caps} solve the task of 3D object classification on point clouds with pruned capsule networks. They compress robust capsule models to deploy them on resource-constrained devices. \textcolor{ForestGreen2}{\citet{Liu_2019_ICCV} propose a new salient property of part-object relationships provided by the CapsNet for salient object detection. The proposed two-stream model requires less computation budgets while obtaining better wholeness and uniformity of the segmented salient object. \citet{disentagled-caps-routing} reduce routing computational complexity with a novel disentangled part-object relational network. Finally, \citet{zhang22} enhances salient object detection by leveraging part-whole hierarchies and contrast cues, integrating attention and background suppression modules for state-of-the-art performance.}

\textcolor{ForestGreen}{Our approach is distinct from prior works. In fact, REM is not intended to be a pruning strategy per se but rather a method to facilitate low-entropy, and thus more interpretable, configurations in the parse trees.} We focus on improving CapsNets part-whole hierarchies extraction from the input images rather than only on their generalization ability. We show that capsules do not need to encompass every detail of the input image to achieve high performance. We illustrate that reasonably sized CapsNets can focus only on the most discriminative object parts that require detection. This achievement is accomplished through regularization and pruning to minimize the entropy of the connections computed by the routing algorithm. 
\section{Proposed method}
\label{sec:method}
The coupling coefficients computed by the routing mechanism model are the part-whole relationships between capsules of two consecutive layers. 
Assigning parts to objects (namely, learning how each object is composed) is challenging. 
One of the main goals of the routing algorithm is to extract a parse tree of these relationships. 
For example, given the $\xi$-th input of class $j$, an ideal parse tree for a capsule $i$ detecting one of the discriminative parts of the entity in the input $\xi$ would ideally lead to
\begin{equation}
    \boldsymbol{c}_{i,:}^{(\xi)} = \mathbb{1}_{y^{(\xi)}},
\end{equation}
where $\mathbb{1}_{y^{(\xi)}}$ is the one-hot encoding for the target class $y^{(\xi)}$ of the $\xi$-th sample. This means that the routing process can carve a parse tree out of the CapsNet, which explains perfectly the relationships between parts and wholes.
One of the problems of this routing procedure is that there is no constraint on how many parse trees there should be. Consequently, a primary capsule can be connected with high probabilities to more than one output capsule. Furthermore, CapsNets have a high number of trainable parameters, and there might be too many active primary capsules, so in this setting, finding the agreements between capsules can be problematic~\cite{renzulli}. 
In this section, we present our techniques first to extract fewer parse trees and then
how to visualize these parse trees. 

\subsection{Routing Entropy Minimization}\label{sec:rem}
\begin{figure}
    \centering
    \includegraphics[width=1\columnwidth]{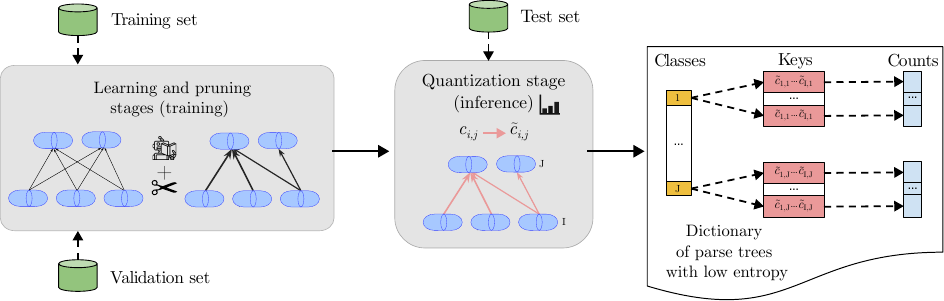}
    \caption{Pipeline\label{fig:pipeline} of REM. After the learning and pruning stages, the coupling coefficients of the CapsNet are quantized, and the obtained parse trees are collected in a dictionary with low entropy.}
\end{figure}

    Here, we present our technique REM. The pipeline of our method is depicted in Figure~\ref{fig:pipeline}. \textcolor{ForestGreen}{It consists of the following steps:
    \begin{enumerate}
        \item we enrich the training process of a CapsNet with pruning to obtain sparse capsule representations (Section~\ref{sec:pruning});
        \item we quantize the coupling coefficients during inference to compute the entropy in the next step (Section~\ref{sec:quantizatino});
        \item we compute a quantitative measure (entropy) to define the cardinality of the extracted parse trees (Section~\ref{sec:entropy}).
    \end{enumerate}}

\subsubsection{Learning and pruning}
\label{sec:pruning}
We train CapsNets via standard back-propagation learning, minimizing some loss functions like margin loss. Our ultimate goal is to assess to what extent a variation of the value of some parameter $\theta$ would affect the error on the network output. In particular, the parameters not affecting the network output can be pushed to zero, meaning we enrich the training stage of CapsNets with pruning. To understand the effect of pruning on the cardinality of the parse trees, we first more formally analyze the distribution of the coupling coefficients
\begin{equation}
    \label{eq::standardcij}
    c_{i,j} = \frac{\exp\left(b_{i,j} + \sum_{r=1}^{r_T} \boldsymbol{u}_j^{(r)} \boldsymbol{u}_{i} \boldsymbol{W}_{i,j}\right)}{\sum_k \exp\left(b_{i,k} + \sum_{r=1}^{r_T} \boldsymbol{u}_k^{(r)} \boldsymbol{u}_i \boldsymbol{W}_{i,k}\right)},
\end{equation}
where $r_T$ indicates the target routing iterations in a CapsNet with two capsule layers. We suppress the indices $\xi$ and $l$ for abuse of notation. Let us evaluate the $c_{i,j}$ over a non-yet trained model: namely we have
\begin{equation}
    \label{eq::uniformcij}
    c_{i,j} \approx \frac{1}{J}\ \forall i,j.
\end{equation}
When updating the parameters, following~\cite{gu}, we have 
\begin{equation}
    \frac{\partial \Ls}{\partial \boldsymbol{W}_{i,j}} = \Bigg[\frac{\partial \Ls}{\partial \boldsymbol{u}_j} \frac{\partial \boldsymbol{u}_j}{\partial \boldsymbol{s}_j} \cdot c_{i,j} + \sum_{m=1}^J \Bigg( \frac{\partial \Ls}{\partial \boldsymbol{u}_m} \frac{\partial \boldsymbol{u}_m}{\partial \boldsymbol{s}_m} \cdot \boldsymbol{\hat{u}}_{i,m}\frac{\partial c_{i,m}}{\partial \boldsymbol{\hat{u}}_{i,j}}\Bigg) \Bigg]\cdot \boldsymbol{u}_i,
\end{equation}
where we can have the gradient for $\boldsymbol{W}_{i,j}\approx 0$ in a potentially-high number of scenarios, despite $c_{i,j}\neq\{0, 1\}$. \textcolor{ForestGreen2}{A step-by-step derivation can be found in~\ref{app:derivation}}. Let us analyze the simple case in which we have perfect outputs, matching the ground truth. Hence, we are close to a local (or potentially the global) minimum of the loss function:
\begin{equation}
    \left\|\frac{\partial \Ls}{\partial \boldsymbol{u}_m} \right\|_2 \approx 0\ \forall m.
\end{equation}
Looking at \Eqref{eq:softmax}, we see that the right class is chosen, but given the squashing function, we have as an explicit constraint that, given the $j$-th class as the target one, we require 
\begin{equation}
    \label{eq::constraint}
    \left\|\boldsymbol{u}_j\right\|_2 \gg \left\|\boldsymbol{u}_m\right\|_2\ \forall m\neq j
\end{equation}    
on the $\boldsymbol{W}_{i,j}$, which can be accomplished in many ways, including:
\begin{itemize}
    \item having sparse activation for the primary capsules $\boldsymbol{u}_i$: in this case, we have constant $\boldsymbol{W}_{i,j}$ (typically associated with no-routing based approaches); however, we need heavier deep neural networks as they have to force sparse signals already at the output of the primary capsules. In this case, the coupling coefficients $c_{i,j}$ are also constant by definition;
    \item having sparse votes $\boldsymbol{\hat{u}}_{i,j}$: this is a combination of having both primary capsules and weights $\boldsymbol{W}_{i,j}$ enforcing sparsity in the votes and the typical scenario with many routing iterations.
\end{itemize}
Having sparse votes, however, does not necessarily result in having sparse coupling coefficients. 
As the votes, $\boldsymbol{\hat{u}}_{i,j}$ are implicitly sparse (yet also disordered, as we are not explicitly imposing any structure in the coupling coefficients distribution), the model is still able to learn, but it finds a typical solution where $c_{i,j}$ are not sparse. However, we would like sparsely distributed, recurrent couplings to the same $j$-th output caps $\boldsymbol{c}_{:,j}$, establishing fewer and more stable relationships between the features extracted at the primary capsule layer.
Hence, we implicitly enforce routing entropy minimization by forcing a sparse and organized structure in the coupling coefficients. Towards this end, our efficient solution is to employ pruning to enforce sparsity in the $\boldsymbol{W}_{i,j}$ representation by compelling a vote between the $i$-th primary capsule and the $j$-th output caps to be exactly zero for any input, according to \Eqref{eq::standardcij}:
\begin{equation}
        c_{i,j} = \frac{1}{\sum_k \exp\left(b_{i,k} + \sum_{r=1}^t \boldsymbol{u}_k^{(r)} \boldsymbol{u}_i \boldsymbol{W}_{i,k}\right)}.
\end{equation}
In this way, having a lower variability in the $c_{i,j}$ values (and hence building more stable relationships between primary and output capsules), straightforwardly, we are also explicitly minimizing the cardinality of the parse trees, namely the entropy of the quantized representations for
the coupling coefficients, that we will define in Section~\ref{sec:entropy}. Note that since we can plug any pruning strategy, the additional training computational cost of REM is the same as the underlying mechanism employed. 

\subsubsection{Quantization} \label{sec:quantizatino}
During the quantization stage, we first compute the \textit{continuous} coupling coefficients $c_{i,j}^{[l](\xi)}$ for each $\xi$-th input example and capsule layer $l$. It should be noted that these are the coupling coefficients obtained after the forward pass of the last routing iteration. Then, we quantize them into $K$ discrete levels through the uniform quantizer $q_K(\cdot)$, obtaining
\begin{equation}
    \tilde{c}_{i,j}^{[l](\xi)} = q_K({c}_{i,j}^{[l](\xi)}).
\end{equation}
We choose the lowest $K$ such that the accuracy does not deteriorate. The quantization stage is performed only during inference, so additional training costs are not required. 

\subsubsection{Entropy} 
\label{sec:entropy}
We want to define a metric that helps us decide if the relationships captured by the routing algorithm resemble a parse tree or not. Towards this end, we can compute the number of occurrences of each coupling sequence to measure the entropy of the whole dictionary. We refer to this entropy as the \textit{simplicity} of the parse tree. In other words, we refer to the number of keys in the dictionary as the number of unique parse trees that can be carved out from the input dataset. 
Given the quantized coupling coefficients of a CapsNet, we can extract the parse tree (and create a dictionary of parse trees) for each class $j$, where each entry is a string composed of the quantization indices of the coupling coefficients $\boldsymbol{\tilde{c}}_{:,j}^{[L-1](\xi)}$. 
Given a dictionary for the coupling coefficients of a CapsNet, we can compute the entropy for each class as 
\begin{equation}
    \label{eq::Hj}
        \mathbb{H}_j = - \sum\limits_{\xi}\big\{ \mathbb{P}\big(\boldsymbol{\tilde{c}}_{:,j}^{[L-1](\xi)} \mid \hat{y}^{(\xi)}\!=\!j\big) \cdot 
        \log_2 \big[\mathbb{P}(\boldsymbol{\tilde{c}}_{:,j}^{[L-1](\xi)} \mid \hat{y}^{(\xi)}\!=\!j)\big]\big\},
\end{equation}
where $\mathbb{P}(\boldsymbol{\tilde{c}}_{:,j}^{[L-1](\xi)} \mid \hat{y}^{(\xi)} = j)$ is the frequency of occurrences of a generic string $\xi$ for each \emph{predicted} class $\hat{y}^{(\xi)}$.
When stacking multiple capsule layers, we apply the quantization stage to each layer and compute the entropy values on the last layer $L$.
Finally, the entropy of a dictionary for a CapsNet on a given dataset is the average of the entropies $\mathbb{H}_j$ of each class
\begin{equation}
    \label{eq::H}
        \mathbb{H} = \frac{1}{J}\sum\limits_{j} \mathbb{H}_j.
\end{equation}
Intuitively, the lower \Eqref{eq::H}, the fewer the number of parse trees carved out from the routing algorithm. 

\subsection{Parse trees visualization}\label{sec:extract_salmap}
As mentioned in Section~\ref{sec:capsnets_fund}, there are three types of capsule layers. Primary and convolutional capsules have a spatial connotation, while fully-connected capsules do not. We describe here two different visualization methods of the extracted parse trees based on the type of the capsules in layer $L-1$. In both scenarios, we follow only the connections starting from the capsule representing the predicted object label; we refer to this method as ``backtracking''. 

\subsubsection{Primary and convolutional capsules}
\begin{figure}
\centering
\includegraphics[width=0.95\columnwidth]{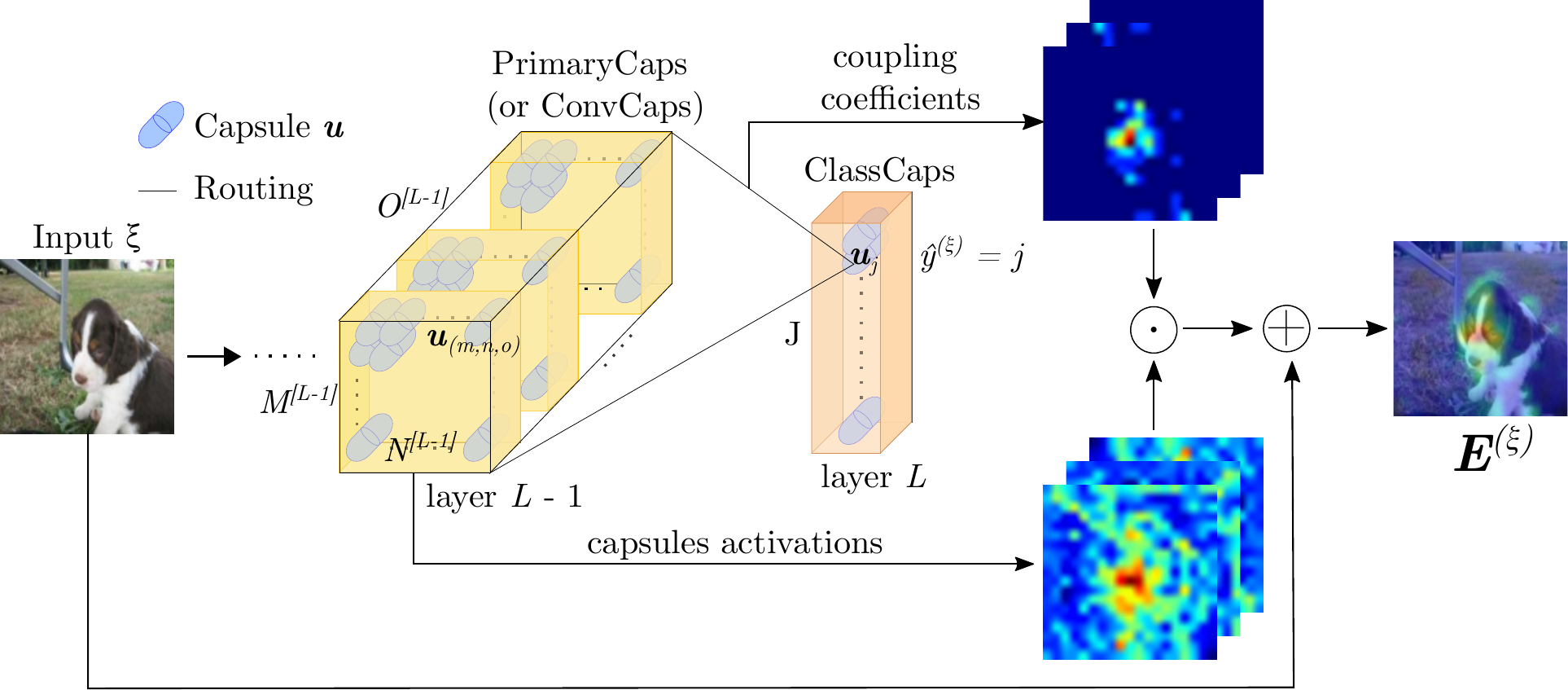}
\caption{\textcolor{black}{Extraction of a saliency map exploiting capsule activations and coupling coefficients. \textcolor{ForestGreen}{This method can be applied when capsules are organized in a $M \times N \times O$ grid structure.}}\label{fig:rem_heatmap_extraction}}
\end{figure}
In Figure~\ref{fig:capsnet_encoder}, primary and convolutional capsules are organized in three dimensions in a $M \times N \times O$ grid structure. Therefore, in this scenario, we exploit the coupling coefficients $c_{i,j}^{[L-1]}$ for the predicted class $j$ as a visual attention built-in explanation to carve out the part structure discovered by a capsule model. We say built-in explanation because we rely only on the forward pass of the network. While Grad-CAM~\cite{gradcam} weighs neuron activations by gradients computed in the backward pass, we weigh each capsule activation by the corresponding coupling coefficient. We refer to $\vu_{(m,n,o)}^{(\xi)}$ to indicate the pose of the capsule in position $(m,n,o)$ for a given input $\xi$. With $\tilde{c}_{(m,n,o),j}^{[L-1](\xi)}$, we refer to the quantized coupling coefficient between a capsule in position $(m,n,o)$ of layer $L-1$ and the predicted capsule-class $j$ of layer $L$  for a given input $\xi$. We denote with $\mE^{(\xi)}$ the saliency map for a given input $\xi$.
We follow \citep{gracapsnets}, where the coupling coefficients of the predicted class $j$ of a trained model for a given input are used as an attention matrix. Unlike \citep{gracapsnets}, we also weight each coupling coefficient by the activity of capsule $i$. 
Therefore, each element of the saliency map is computed as 
\begin{equation}
    E_{(m,n)}^{(\xi)} = \frac{1}{O}\sum\limits_{o}(\|\vu_{(m,n,o)}^{[L-1](\xi)}\| \cdot \tilde{c}_{(m,n,o),j}^{[L-1](\xi)}).
\end{equation}
Then we upsampled the saliency map to the input size with the bilinear method.
Figure~\ref{fig:rem_heatmap_extraction} depicts our method's visualization of extracting a saliency map from an input image given a CapsNet model. 

\subsubsection{Fully-connected capsules}
CapsNets can also be organized in fully connected capsule layers. The saliency map method can not be employed in this scenario, as capsules in layer $L-1$ do not have a spatial connotation. Therefore, similarly to~\citet{mitterreiter2023capsule}, we propose a different visualization of the parse tree. Each capsule is a node where the color is its activation probability, and the strength of the connection is depicted with fuzzy edges of different strokes (as in the toy example in Figure~\ref{fig:parse_tree}). We also denote with $deg^-(\vu_j)$ the in-degree of a capsule $j$ in layer $l$, which is the number of coupling coefficients starting from capsules $i$ in layer $l-1$ above some threshold. We only show the connections from capsule $j$ to capsules $k$ in layer $l+1$ if $deg^-(\vu_j)$ is at least 1. \textcolor{ForestGreen2}{Note that for high-resolution datasets with many classes, saliency map visualizations (for capsule layers with spatial structure) are preferable over node-based parse tree visualizations, as they provide a clearer and more interpretable representation of capsule activations.}

\section{Experiments and results}
\label{sec:results}
This section reports the experiments and results we performed to test REM. 
We will refer to CapsNet+REM as a trained CapsNet where we employ both pruning and quantization stages, and CapsNet+Q when only the quantization is applied. \textcolor{ForestGreen}{The code can be found at \url{https://github.com/EIDOSLAB/REM}.}
\subsection{Datasets setup and metrics}
\label{sec:datasets-setup-metrics}
We train and test REM on many different datasets: MNIST~\citep{mnist} and Fashion-MNIST~\citep{fashion-mnist}, 28$\times$28 grayscale images (10 classes); SVHN~\citep{SVHN}, 32$\times$32 RGB images (10 classes); CIFAR-10~\citep{CIFAR10}, 32$\times$32 RGB images (10 classes); Dogs vs. Cats~\citep{dogs-vs-cats}, images resized to 128$\times$128; Imagenette~\citep{imagewang}, a subset of 10 classes from the Imagenet dataset resized to 128$\times$128. We used 10\% of the training sets as validation sets. 
We further test the robustness of our technique to affine transformations and to novel viewpoints.
For affine transformations, we use expanded MNIST training and validation sets (40×40 padded and translated MNIST images) and the affNIST~\citep{affNIST} test set, in which each example is an MNIST digit with a random small affine transformation.
For novel viewpoints such as novel azimuths and elevation, we use smallNORB~\citep{smallNORB}. We train all models on 1/3 of the training data with azimuths of~$~{\{0, 20, 40, 300, 320, 340\}}$ degrees and test them on 2/3 of the test data with remaining azimuths never seen during training. We also trained models on 1/3 of the training data with elevations of~$~{\{30, 35, 40\}}$ degrees from the horizontal. We tested on 2/3 of the test data with the remaining elevations. 

We conducted our experiments with five random seeds. We report the classification accuracy (\%) and entropy (averages and standard deviations), the \textit{sparsity} (percentage of pruned parameters, median), and the number of keys in the dictionary (median).

\subsection{Capsule architectures}
We also test the efficacy of REM to other variants of capsule models, including different architectures, routing algorithms, and the number of trainable parameters.  We conducted experiments applying our technique to DR-CapsNets~\cite{hinton-dynamic}, $\gamma$-CapsNets~\citep{peer2019gammacapsules},
DeepCaps~\citep{deepcaps} and Eff-CapsNets~\citep{mazzia2021efficient}. All models employed in this work use the same architectures (number of layers, capsule dimensions, number of routing iterations, etc.) presented
in the original papers. We used a custom architecture called Eff-ConvCapsNet for higher-resolution images such as Dogs vs. Cats and Imagenette. This model has three capsule layers (primary, convolutional and fully-connected) and shares a backbone network similar to Eff-CapsNets.  

We apply iterative magnitude pruning~\citep{lobster} as the main pruning strategy since it can prune many parameters without performance loss, and it can be easily coded into the training process.
More details on the architectures, training process, choice of quantization levels, and the decoder part are reported in~\ref{sec:model_architectures}, ~\ref{sec:training}, \ref{sec:choice_quant_levels} and \ref{subsec:decoder}, respectively. The experiments were run on an NVIDIA Ampere A40 equipped with 48GB RAM, and the code uses PyTorch 1.12.

\subsection{Experiments}
\label{sec:exp}

\subsubsection{Generalization ability} 
As shown in Table \ref{tab:results-otherdatasets}, a CapsNet+REM has a high percentage of pruned parameters with a minimal performance loss. So, this confirms our hypothesis that CapsNets are over-parametrized.
We also report the entropy of the dictionary of the last routing layer for the quantized models. Since vanilla CapsNets are not pruned or quantized, we can not compute the entropy, and they are not sparse (so sparsity is always 0, the same for CapsNets+Q). The entropy is successfully lower for all datasets when REM is applied to all architectures, even with fewer parameters than CapsNets, such as Eff-CapsNets. 
Compared to other models, Eff-CapsNets have lower entropy since, in the original implementation, the kernel size for the PrimaryCaps has the same dimension as the backbone output. Therefore, in this case, PrimaryCaps has no spatial resolution. Furthermore, we noticed that apart from simple datasets such as MNIST or Fashion-MNIST, the coupling coefficient distributions collapse to similar values, and the non-iterative self-attention mechanism does not build a proper parse tree.
Extended results are reported in ~\ref{sec:other_distr_tables}.
\begin{table}[h]
\footnotesize
\centering
\resizebox{\textwidth}{!}{
\begin{tabular}{cc  ccc  ccc  ccc}
\toprule


\multirow{2}{*}{\textbf{Dataset}} & \multirow{2}{*}{\textbf{Architecture}} &  \multicolumn{1}{c}{\bf Vanilla} &\multicolumn{2}{c}{\bf Quantized (Q)} & \multicolumn{3}{c}{\bf REM}\\

&& \textbf{Accuracy} & \textbf{Accuracy} & \textbf{Entropy} & \textbf{Accuracy} & \textbf{Entropy} & \textbf{Sparsity}\\
\midrule

\multirow{4}{*}{F-MNIST} & 
DR-CapsNet              & $92.76\pm_{0.21}$ & $92.46\pm_{0.23}$ & $8.64\pm_{1.15}$ & $92.62\pm_{0.14}$ & $4.80\pm_{1.70}$ & $80.71$\\
& $\gamma$-CapsNet      & $92.59\pm_{1.14}$ & $92.43\pm_{1.10}$ & $3.98\pm_{0.76}$ & $93.01\pm_{1.06}$ & $1.45\pm_{0.68}$ & $87.07$ \\
& DeepCaps              & $92.36\pm_{0.27}$ & $92.33\pm_{0.24}$ & $7.15\pm_{1.33}$ & \cellcolor[HTML]{FFF8AD}$94.61\pm_{0.06}$ & $6.08\pm_{1.29}$ & $83.29$\\
& Eff-CapsNet           & $93.31\pm_{0.31}$ & $93.22\pm_{0.24}$ & $3.88\pm_{1.10}$ & $92.98\pm_{0.42}$ & \cellcolor{LightGreen} $1.10\pm_{0.48}$ & $63.29$ \\ \midrule

\multirow{4}{*}{SVHN} & 
DR-CapsNet              & $93.30\pm_{0.41}$ & $92.20\pm_{0.20}$ & $7.13\pm_{1.15}$ & $91.71\pm_{0.53}$ &  $5.23\pm_{0.71}$ & $74.40$\\
& $\gamma$-CapsNet      & $89.02\pm_{0.12}$ & $87.42\pm_{1.22}$ & $7.15\pm_{0.86}$ & $88.36\pm_{0.21}$ & $5.65\pm_{1.22}$ & $73.89$ \\
& DeepCaps              & $93.32\pm_{0.36}$ & $93.20\pm_{0.36}$ & $11.06\pm_{0.58}$ & $93.06\pm_{0.24}$ & $3.97\pm_{1.50}$ & $80.50$\\
& Eff-CapsNet           & \cellcolor[HTML]{FFF8AD}$93.64\pm_{0.11}$ & $93.62\pm_{0.03}$ & $0.53\pm_{0.59}$ & $93.12\pm_{0.02}$ & \cellcolor{LightGreen} $0.24\pm_{0.41}$ & $47.80$ \\ \midrule

\multirow{4}{*}{CIFAR-10} & 
DR-CapsNet              & $79.93\pm_{0.28}$ & $78.42\pm_{2.06}$ & $6.26\pm_{0.61}$ & $79.25\pm_{0.59}$ & $4.15\pm_{0.62}$ & $81.17$\\
& $\gamma$-CapsNet      & $74.02\pm_{0.31}$ & $73.08\pm_{0.48}$ & $3.67\pm_{0.70}$ & $74.89\pm_{0.23}$ & $3.22\pm_{0.66}$ & $74.89$\\
& DeepCaps              & \cellcolor[HTML]{FFF8AD}$90.80\pm_{0.13}$ & $90.47\pm_{0.10}$ & $8.99\pm_{0.52}$ & $90.35\pm_{0.13}$ & $7.07\pm_{1.01}$ & $46.83$ \\
& Eff-CapsNet           & $81.53\pm_{0.61}$ & $81.51\pm_{0.52}$ & $0.25\pm_{0.35}$ & $81.49\pm_{0.47}$ & \cellcolor{LightGreen} $0.01\pm_{0.02}$ & $53.79$ \\
\midrule

\multirow{1}{*}{Dogs vs. Cats} & 
Eff-ConvCapsNet              & $97.74\pm_{0.37}$ & $99.20\pm_{0.25}$ & $4.05\pm_{0.28}$ & \cellcolor[HTML]{FFF8AD}$99.10\pm_{0.56}$ & \cellcolor{LightGreen} $2.03\pm_{0.73}$ & $71.92$\\
\midrule

\multirow{1}{*}{Imagenette} & 
Eff-ConvCapsNet              & \cellcolor[HTML]{FFF8AD}$83.11\pm_{0.18}$ & $82.98\pm_{0.66}$ & $7.63\pm_{0.51}$ & $83.02\pm_{0.78}$ & \cellcolor{LightGreen} $5.85\pm_{0.38}$ & $63.52$\\
\bottomrule
\end{tabular}}
\caption{\centering\label{tab:results-otherdatasets}Accuracy (\%), entropy and sparsity (\%) on Fashion-MNIST, SVHN, CIFAR-10, Dogs vs. Cats and Imagenette (test set). \textcolor{ForestGreen2}{We highlight in \colorbox{LightGreen}{green} the lowest entropy and in \colorbox{Yellow}{yellow} the best accuracy for each dataset.}}
\end{table}
\subsubsection{Robustness to affine transformations}
To test the robustness to affine transformations of CapsNets+REM, we trained on expanded MNIST and tested on affNIST. We tested an under-trained CapsNet with early stopping, which achieved 99.22\% accuracy on the expanded MNIST test set as in ~\citep{hinton-dynamic, gu}. We also trained these models until convergence. We can see in Table~\ref{tab:results-affnist} that the under-trained network entropies are high. Instead, a well-trained CapsNet+REM can be robust to affine transformations and have low entropy.
\begin{table}[ht]
    \footnotesize
    \centering
    \begin{tabular}{ccccc}
        \toprule & \textbf{expanded MNIST} & \textbf{affNIST} & \textbf{affNIST} & \textbf{affNIST}  \\ 
        \textbf{Model} & \textbf{Accuracy} & \textbf{Accuracy} & \textbf{\begin{tabular}[c]{@{}c@{}}Entropy \end{tabular}} & \textbf{Sparsity} \\ \midrule
        DR-CapsNet+Q   & $99.22$           & $77.93\pm_{0.55}$ & $8.64\pm_{1.15}$ & $0$      \\
        DR-CapsNet+REM & $99.22$           & \cellcolor{Yellow}$81.81\pm_{0.81}$ & \cellcolor{LightGreen}$8.45\pm_{1.10}$ & $71.26$  \\ \midrule
        DR-CapsNet+Q   & $99.36\pm_{0.05}$ & $83.14\pm_{0.24}$ & $8.45\pm_{0.99}$ & $0$      \\
        DR-CapsNet+REM & $99.48\pm_{0.02}$ & \cellcolor{Yellow}$85.23\pm_{0.11}$ & \cellcolor{LightGreen}$5.93\pm_{1.39}$ & $87.32$  \\ \midrule
    \end{tabular}
    \caption{\centering\label{tab:results-affnist}{Accuracy (\%), entropy and sparsity (\%) on affNIST test set for under-trained and well-trained models. \textcolor{ForestGreen2}{We highlight in \colorbox{LightGreen}{green} the lowest entropy and in \colorbox{Yellow}{yellow} the best accuracy.}}}
\end{table}

\subsubsection{Robustness to novel viewpoints}
CapsNets are well known for their generalization ability to novel viewpoints~\citep{hinton-dynamic, hinton-em}.
We conducted further experiments on the smallNORB dataset to test the robustness of novel viewpoints of our technique following the experimental protocol of \citep{self-routing, hinton-em} described in Section~\ref{sec:datasets-setup-metrics}.
We employed Eff-CapsNets, as they are the state-of-the-art models on this dataset with few trainable parameters.
We used $K=11$ quantization levels for Eff-CapsNets+Q and Eff-CapsNets+REM.
In Table~\ref{tab:novelviewpoints}, we can see that Eff-CapsNets+REM are indeed robust to novel viewpoints with low entropy. Notice that even if our work does not target state-of-the-art generalization, with REM, we manage to achieve a maximum value of accuracy of 86.99\% with only 27k trainable parameters on novel elevations (98.43\% for familiar elevations) and 90.37\% with only 37k trainable parameters on novel azimuths (97.53\% for familiar azimuths).  See ~\ref{sec:novel_viewpoints} for additional results with different networks.
\begin{table}[h]
\footnotesize
\centering
\begin{tabular}{lccccc}
\toprule
\multicolumn{1}{c}{\multirow{2}{*}{\textbf{Model}}} & \multicolumn{2}{c}{\textbf{Familiar}}                               & \multicolumn{2}{c}{\textbf{Novel}}                                  & \multicolumn{1}{l}{\multirow{2}{*}{\textbf{Sparsity}}} \\ 
                        & \textbf{Accuracy} & \multicolumn{1}{l}{\textbf{Entropy}} & \multicolumn{1}{l}{\textbf{Accuracy}} & \multicolumn{1}{l}{\textbf{Entropy}} & \multicolumn{1}{l}{}                          \\ \midrule
\textcolor{black}{Eff-CapsNet+Q ($\phi$)} & $97.81\pm_{0.45}$ & $1.10\pm_{0.42}$ & $85.03\pm_{1.01}$ & $1.33\pm_{0.53}$ & $0$ \\ 
\textcolor{black}{Eff-CapsNet+REM ($\phi$)} & \cellcolor{Yellow}$98.08\pm_{0.25}$ & \cellcolor{LightGreen}$0.26\pm_{0.14}$ & \cellcolor{Yellow}$86.50\pm_{0.31}$ & \cellcolor{LightGreen}$0.25\pm_{0.11}$ & $75.09$ \\ \midrule
\textcolor{black}{Eff-CapsNet+Q ($\psi$)} & $97.07\pm_{0.52}$ & $1.69\pm_{0.28}$ & $87.95\pm_{1.67}$ & $1.71\pm_{0.34}$ & $0$ \\ 
\textcolor{black}{Eff-CapsNet+REM ($\psi$)} & \cellcolor{Yellow}$97.31\pm_{0.31}$ & \cellcolor{LightGreen}$0.38\pm_{0.15}$ & \cellcolor{Yellow}$88.48\pm_{1.38}$ & \cellcolor{LightGreen}$0.33\pm_{0.12}$ & $68.85$\\ \bottomrule                      
\end{tabular}
\caption{\centering\label{tab:novelviewpoints}\textcolor{black}{Accuracy (\%), entropy and sparsity (\%) on the smallNORB test set on familiar and novel viewpoints (elevations $\phi$ and azimuths $\psi$) seen and unseen during training respectively. \textcolor{ForestGreen2}{We highlight in \colorbox{LightGreen}{green} the lowest entropy and in \colorbox{Yellow}{yellow} the best accuracy for each dataset.}}}
\end{table}

\subsubsection{Improved visualizations with REM}\label{sec:vis-rem}
\begin{figure}[h]
\centering
\begin{subfigure}{1\columnwidth}
  \centering
  \includegraphics[width=0.85\textwidth]{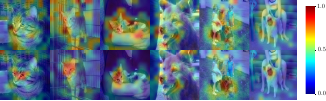}
  \caption{\centering{Dogs vs. Cats.}}
  \label{fig:attention_matr_dogscats}
\end{subfigure}\\
\begin{subfigure}{1\columnwidth}
  \centering
  \includegraphics[width=0.85\textwidth]{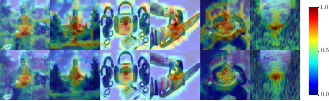}
  \caption{\centering{Imagenette.}}
  \label{fig:attention_matr_imagenette}
\end{subfigure}%
\caption{Saliency maps for Eff-ConvCapsNets without (first rows) and with REM (second rows)}
\label{fig:saliency_maps}
\end{figure}

\begin{figure}[h]
\centering
\includegraphics[width=0.9\columnwidth]{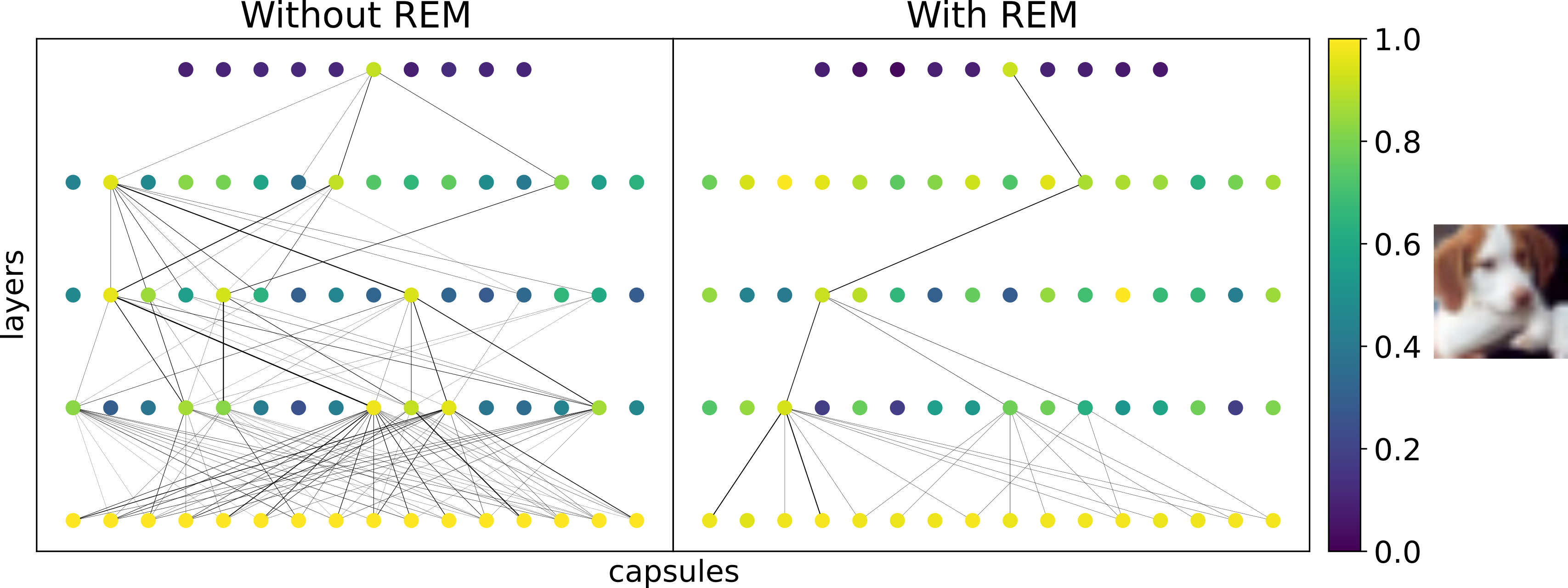}
\caption{Parse trees of a fully connected multilayer DR-CapsNet and DR-CapsNet+REM for a CIFAR-10 test image. \label{fig:CIFAR-10_capsnet_vs_remcaps}}
\end{figure}

Figures~\ref{fig:attention_matr_dogscats} and  \ref{fig:attention_matr_imagenette} show the saliency maps overlayed on Dogs vs. Cats and Imagenette images. We chose these datasets for the saliency maps visualization method since the images come with higher resolution and more variation than the other datasets.
When REM is applied (second rows), we can see that the network predominantly concentrates on the object of interest, dismissing extraneous background noise or irrelevant objects. This characteristic of REM leads to the extraction of more concise part-whole hierarchies. As illustrated in Figure~\ref{fig:attention_matr_dogscats}, only the discriminative features specific to a cat or a dog are prominently highlighted in red. In contrast, features corresponding to people or cages play a diminished role in the classification process. Similarly, in Figure~\ref{fig:attention_matr_imagenette}, the model focuses mostly on the tench, the cassette player, the chain saw, the French horn, and the golf ball, disregarding trees, headphones, hands, the man, and grass.
Figure~\ref{fig:CIFAR-10_capsnet_vs_remcaps} shows the parse trees extracted from a fully-connected DR-CapsNet for a CIFAR-10 image of the test set. 
DR-CapsNet (left) and DR-CapsNet+REM (right) comprise a primary capsule layer, three hidden layers, and one last layer for the object labels. We only employed 16 capsules for the primary and hidden layers for more precise visualization. We can see that fewer capsules are part of the parse tree extracted with REM: the network relies on fewer object parts to detect the dog. Additional visualizations are reported in ~\ref{sec:other_visualizations}.

\subsection{Ablation study}
\label{sec:abl}
To assess our technique, we analyze the benefits of REM on the MNIST dataset in depth. Despite its outdatedness, MNIST remains an omnipresent benchmark for CapsNets~\citep{gu, sparse-caps, VAEs_capsules, mitterreiter2023capsule}.

\begin{figure}
\centering
\includegraphics[width=0.8\columnwidth]{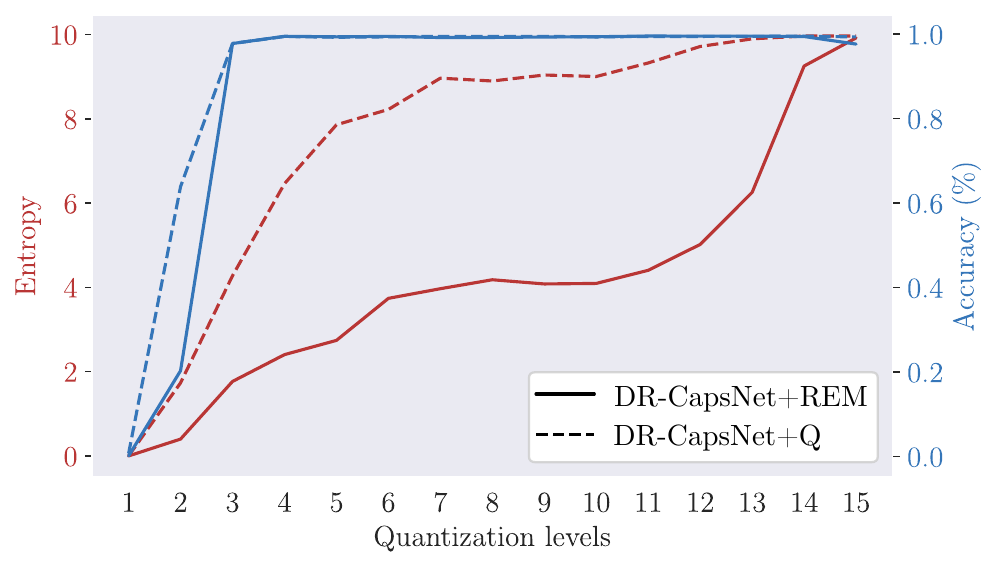}
\caption{\textcolor{ForestGreen2}{Entropy and accuracy values for DR-CapsNet+Q and DR-CapsNet+REM with different quantization levels on MNIST (test set).}\label{fig:quantization levels}}
\end{figure}

\textcolor{ForestGreen2}{\textbf{Choice of quantization levels}.
As we can see in Figure~\ref{fig:quantization levels}, the choice of the number of quantization levels $K$ for the coupling coefficients computed by a routing algorithm of a DR-CapsNet affects the performance of the network. We select the value for $K$ that achieves the best accuracy value with relatively low entropy. In this case, when $K$=11, DR-CapsNet+Q achieves $99.47\%$ accuracy and $9.32$ entropy, while DR-CapsNet+REM achieves $99.57\%$ accuracy and $4.40$ entropy. For further details of the choice of $K$ on other architectures and datasets, please see~\ref{sec:choice_quant_levels}.}

\textbf{Entropy at different epochs}. On a given dataset, we target a model with high generalization but low entropy, namely a low number of extracted parse trees.
\begin{figure}
  \centering
  \includegraphics[width=0.8\columnwidth]{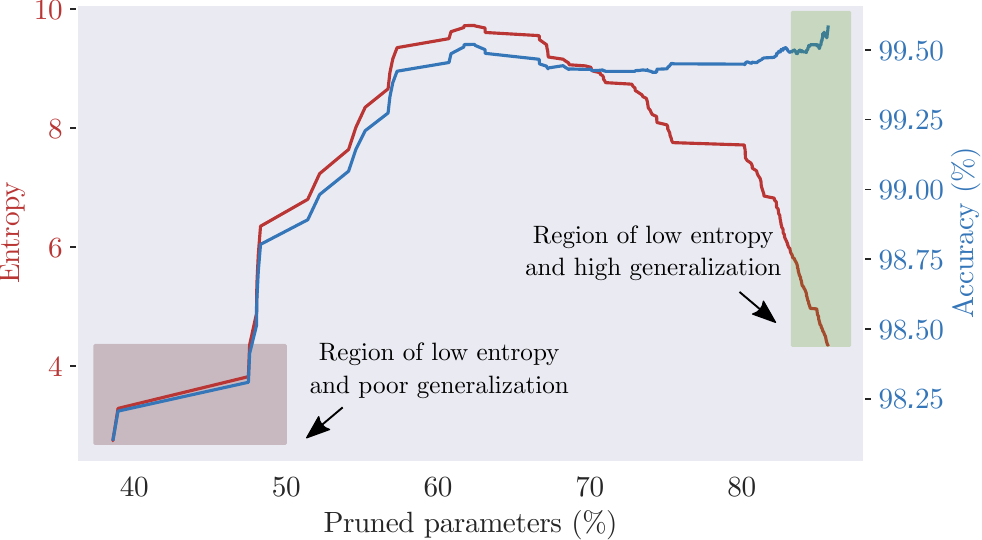}
  \caption{Entropy (red line) and accuracy (blue line) of DR-CapsNe+REM at different epochs (with iterative pruning) on MNIST (test set).}
  \label{fig:acc_vs_pp}
\end{figure}
Figure~\ref{fig:acc_vs_pp} shows how the entropy and classification accuracy change as the sparsity increases during training with $K = 11$. At the beginning of the training stage, we can see that the entropy is low (1.83) because the routing algorithm has not yet learned to correctly discriminate the relationships between the capsules (97.53\% accuracy). This effect is almost the same when we train a DR-CapsNet with $t=1$ as \citet{gu}, where its entropy is zero, but capsules are uniformly coupled. However, at the end of the training process, we can get a model trained with REM with higher performances (99.60\% accuracy) and still low entropy (4.31). 

\begin{figure}
 \centering
 \includegraphics[width=0.9\columnwidth] {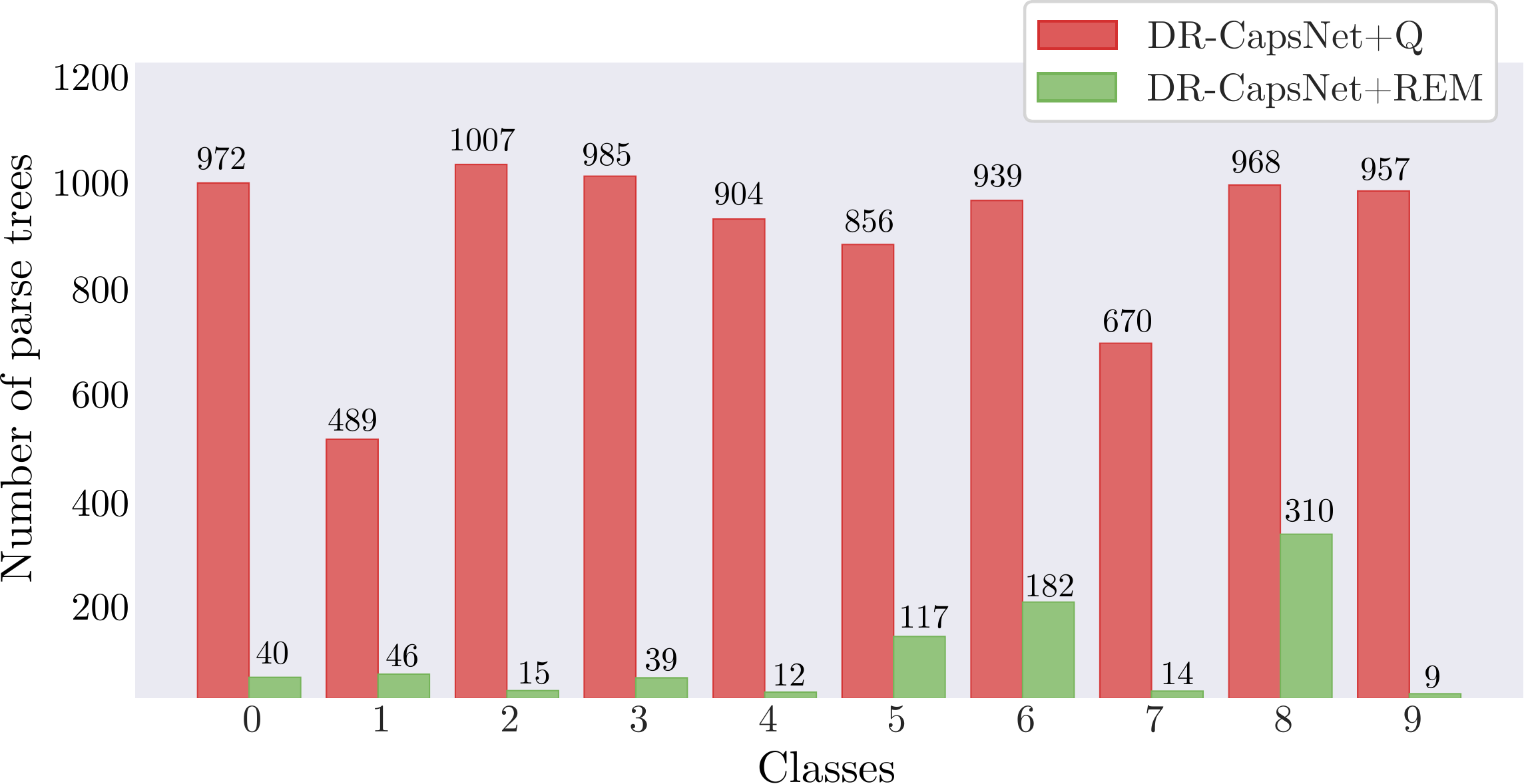}
 \caption{\centering\label{fig:mnist-dictionary-test}Number of keys for each class on MNIST (test set).}
 \end{figure}

\textbf{Number of parse trees.} Figure~\ref{fig:mnist-dictionary-test} shows the number of intra-class parse trees (collected in a dictionary) for DR-CapsNet+REM and DR-CapsNet+Q. We can see that the number of keys in the dictionary for DR-CapsNet+REM is lower than the one for DR-CapsNet+Q for each class. Also, the entropy measure for DR-CapsNet+REM is lower than DR-CapsNet+Q; namely, REM has successfully extracted fewer parse trees on the MNIST test set.

\begin{figure}
 \centering
 \includegraphics[width=0.9\columnwidth]{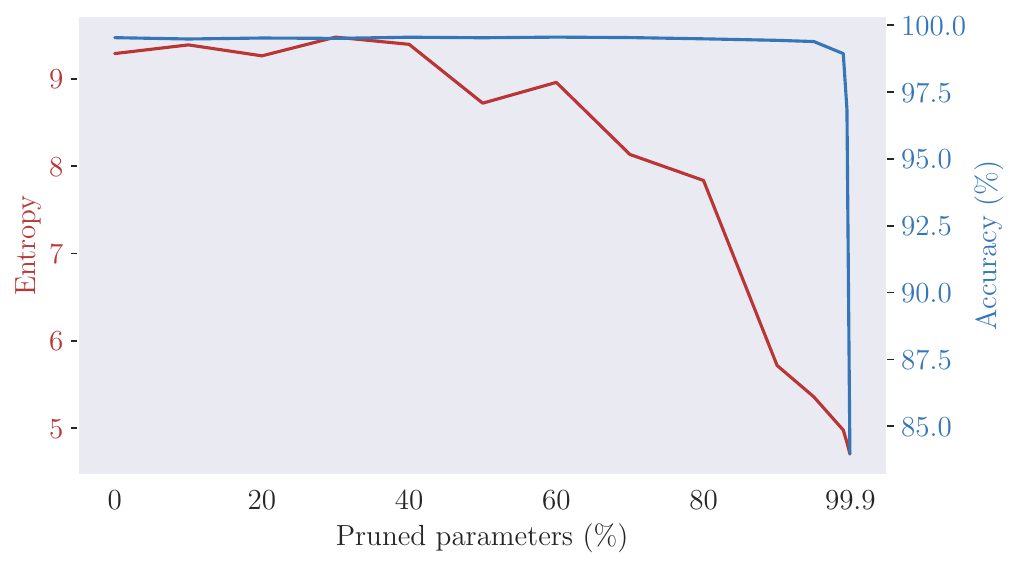}
 \caption{\centering\label{fig:snip}Entropy (red line) and accuracy (blue line) of DR-CapsNe+REM at different epochs (with single-shot pruning) on MNIST (test set).}
 \end{figure}

\textbf{Different pruning strategies.}
With REM, we can plug any pruning strategy. However, our experiments used iterative magnitude pruning, which gradually prunes neural network weights during training, instead of single-shot pruning, which prunes parameters only at initialization. In fact, in Figure~\ref{fig:snip}, we can see that employing a single-shot pruning strategy such as SNIP~\citep{lee2018snip} leads to low entropy but, compared to Figure~\ref{fig:acc_vs_pp}, the performance starts to deteriorate at lower pruning ratios. 
Therefore, due to better performance and its non-intrusiveness in the complex and delicate routing mechanism for CapsNets, iterative magnitude pruning~\citep{lobster} resulted in a suitable choice to enforce REM.

\section{Conclusion}
\label{sec:conlcusion}
\textcolor{ForestGreen}{In this paper, we highlighted the challenges posed by current overparametrized capsule networks in handling irrelevant entities and noisy backgrounds, and how our REM technique addresses these issues by focusing on more relevant features, thus paving the way for more efficient and scalable capsule networks.}
REM drives the model parameters distribution toward low entropy configurations. With REM, capsule networks do not need to model irrelevant objects or noise in the images to achieve high generalization. Therefore, even a reasonable-sized capsule network can now learn more succinct features. Novel visualization methods for capsule networks show which features play a major role in the classification process. 
 We first showed how to extract the parse tree of a capsule network by discretizing its connections and then collecting the possible parse trees in associative arrays.
Capsule networks show high entropy in the parse tree structures, as an explicit prior on the distribution of coupling coefficients is missing. Indeed, the number of intra-class generated parse trees is relatively high. We showed how pruning naturally reduces such entropy and the cardinality over the possible parse trees. Extensive experiments on different datasets and architectures confirm the effectiveness of the proposed approach. Furthermore, we empirically observe that with REM, models remain robust to affine transformations and novel viewpoints. 
Recent works show that capsule networks do not scale~\citep{mitterreiter2023capsule} and suffer vanishing activations when adding many capsule layers~\citep{everett2023vanishing}, leading to poor performance. Therefore, carving out an ideal parse tree from complex images with perfectly interpretable capsules is not possible with capsule networks, yet. However, REM opens research pathways toward the distillation of parse trees and model interpretability, including the design of a pruning technique specifically designed for capsules. 

\textcolor{ForestGreen}{\textbf{Limitations.}
The main limitation of our work arises when a decoder network is attached to a CapsNet. Since the network needs to reconstruct the input images, a fewer number of parameters is removed and the entropy is not decreasing as in the scenario with only the encoder part. Please see ~\ref{subsec:decoder} for a more detailed explanation and results. Furthermore, CapsNets have not yet seen widespread adoption in the industry or research community compared to traditional CNNs or ViTs. The lack of standardized architectures and pre-trained models (for example, on ImageNet) might be a contributing factor. Several obstacles that still need to be addressed are the intensive computational complexity, training instability when stacking more capsule layers, and lack of proper interpretability, despite initial promising results. Addressing these challenges is key to making CapsNets more practical and widely adopted.}

\bibliographystyle{elsarticle-num-names} 
\bibliography{main}







\newpage
\appendix
{\color{ForestGreen}\section{Notation}\label{sec:notation}
We provide in Table~\ref{tab:notation} the notation used in this work.
\begin{table}[H]
\footnotesize
    \centering
    \begin{tabularx}{\textwidth}{lX}
        \hline
        \textbf{Symbol} & \textbf{Description} \\
        \hline
        $y^{(\xi)}$ & The correct label for the $\xi^{th}$ example \\
        $\hat{y}^{(\xi)}$ & The predicted label for the $\xi^{th}$ example \\
        $\Omega^{[l]}$ & Set of capsules in layer $l \in \{1,2,...,L\}$ \\
        $|\Omega^{[1]}|$ & Number of primary capsules \\
        $|\Omega^{[L]}|$ & Number of class capsules \\
        $I$ & Number of primary capsules (when $L=2$) \\
        $J$ & Number of class capsules (when $L=2$) \\
        $\vs^{[l]}_{j} \in \mathbb{R}^{D^{[l]}}$ & Unnormalized $D^{[l]}$-dimensional feature vector of a capsule $j$ in layer $l$ \\
        $\vu^{[l]}_{j} \in \mathbb{R}^{D^{[l]}}$ & Normalized $D^{[l]}$-dimensional feature vector of a capsule $j$ in layer $l$ \\
        $\hat{\vu}^{[l+1]}_{i,j} \in \mathbb{R}^{D^{[l+1]}}$ & Vote/Prediction of capsule $i$ in layer $l$ for capsule $j$ in layer $l+1$ \\
        $\mW^{[l]}_{i,j} \in \mathbb{R}^{D^{[l]} \times D^{[l+1]}}$ & Trainable weight matrix between capsule $i$ in layer $l$ and capsule $j$ in layer $l+1$ \\
        $b^{[l]}_{i,j}$ & Log prior probability that capsule $i$ in layer $l$ should be coupled to capsule $j$ in layer $l+1$ \\
        $c^{[l]}_{i,j}$ & Generic coupling coefficient between capsule $i$ in layer $l$ and capsule $j$ in layer $l+1$ \\
        $c^{[l](\xi)}_{i,j}$ & Coupling coefficient for the $\xi^{th}$ example between capsule $i$ in layer $l$ and capsule $j$ in layer $l+1$ \\
        $r^{[l]}$ & Number of routing iterations between capsule layers $l$ and $l+1$ \\
        $K$ & Number of quantization levels \\
        $\tilde{c}^{[l]}_{i,j}$ & Quantized coupling coefficient between capsule $i$ in layer $l$ and capsule $j$ in layer $l+1$ \\
        $\mathbb{H}_j$ & Entropy of class j \\
        $\mathbf{E}^{(\xi)}$ & Saliency matrix for the $\xi^{th}$ example\\
        \hline
    \end{tabularx}
    \caption{\textcolor{ForestGreen}{List of symbols and their definitions used in the paper.}}
    \label{tab:notation}
\end{table}}

\section{Experiments details}\label{sec:exp-details}
In this section, we provide the technical details of our experiments, including optimizers, hyperparameter values, and architecture configurations. 

\subsection{Model architectures}\label{sec:model_architectures}
All models employed in this work were tested using the same architectures (number of layers, capsule dimensions, number of routing iterations, etc.) presented in the original papers. Therefore, for DR-CapsNets, $\gamma$-CapsNets, DeepCaps and Eff-CapsNets we used the original architecture configurations.

Eff-ConvCapsNet consists of three capsule layers stacked on top of a backbone network similar to Eff-CapsNets. We use the notation Conv(\textit{output-channels}, \textit{kernel}, \textit{stride}) for standard convolutional layers. For convolutional capsule layers, the notation is ConvCaps(\textit{output-capsule-dimension}, \textit{output-capsule-types}, \textit{kernel}, \textit{stride}). For fully-connected capsule layers, we use FcCaps(\textit{output-capsule-dimension}, \textit{output-capsule-types}). The backbone network utilizes four standard convolutional layers Conv(32,5,2), Conv(64,3,1), Conv(64,3,1), and Conv(256,3,2), followed by a fifth depthwise convolutional layer Conv(256,3,1) to extract 16 primary capsule types of dimension 16. Then, a convolutional capsule layer ConvCaps(16,16,3,2) is stacked on top of primary capsules. Finally, a fully-connected capsule layer FcCaps(16, 10) (FcCaps(16, 2) for Dogs vs. Cats) with shared transformation matrices is stacked to output the class capsules. We used three iterations of the dynamic routing method described in Section~\ref{sec:capsnets_fund}.

To visualize the parse trees in multilayer DR-CapsNets, we stacked five capsule layers, 16 capsules for the primary capsule layer, 16 capsules for the three hidden layers and ten capsules for the last layer. Each capsule is composed by 8 neurons.

\subsection{Training}\label{sec:training}
For models without REM, we take the checkpoint that achieved the lowest loss on the validation set, while for models with REM we take the model on the last epoch. We checked the loss on the validation set and used an early-stop of $200$ epochs. The models were trained on batches of size 128 using Adam optimizer with its PyTorch~1.12 default parameters, including an exponentially decaying learning rate factor of $0.99$.

\subsection{Choice of quantization levels}\label{sec:choice_quant_levels}
The routing algorithms used in the models employed in this paper are performed between two consecutive capsule layers. When stacking multiple capsule layers, for example, using $\gamma$-CapsNets and DeepCaps, we apply the quantization stage to each layer and compute the entropy values on the last layer. For each capsule layer, we chose the lowest $K$ such that the accuracy does not deteriorate.
 For $\gamma$-CapsNets+Q and $\gamma$-CapsNets+REM we found $K=11$ for the first two capsule layers and $K=6$ for the last two. For DeepCaps+Q and DeepCaps+REM we used $K=11$ for all the capsule layers where the number of routing iterations is greater than one. For Eff-CapsNets+Q and Eff-CapsNets+REM we used $K=11$ on smallNORB. For Eff-ConvCapsNets+Q and Eff-ConvCapsNets+REM we used $K=21$ on Dogs vs. Cats and Imagenette.

As regards DR-CapsNets+Q and DR-CapsNets+REM, on MNIST, Fashion-MNIST, CIFAR-10, and affNIST, we found $K=11$ for the quantizer, while for Tiny ImageNet we found $K=129$ quantization levels.

\section{Additional and extended results}\label{sec:additional_results}
In this section, we provide additional and extended results for MNIST, Fashion-MNIST, SVHN, CIFAR-10, smallNORB, affNIST, and Tiny ImageNet, including distributions of the coupling coefficients. We also provide additional visualizations employing the dictionary built with our technique REM to understand better the impact of having fewer parse trees with stronger connections.

\subsection{Robustness to novel viewpoints}\label{sec:novel_viewpoints}
CapsNets are well known for their ability to generalize novel viewpoints.
We conducted further experiments on the smallNORB dataset to test the robustness of novel viewpoints of our technique following the experimental protocol described in Section~\ref{sec:datasets-setup-metrics}).
In Table~\ref{tab:novelviewpoints_other}, we can see the effectiveness of REM also on other networks than Eff-CapsNets. 
\begin{table}[h]
\footnotesize
\centering
\begin{tabular}{lccccc}
\toprule
\multicolumn{1}{c}{\multirow{2}{*}{\textbf{Model}}} & \multicolumn{2}{c}{\textbf{Familiar}}                               & \multicolumn{2}{c}{\textbf{Novel}}                                  & \multicolumn{1}{l}{\multirow{2}{*}{\textbf{Sparsity}}} \\ 
                        & \textbf{Accuracy} & \multicolumn{1}{l}{\textbf{Entropy}} & \multicolumn{1}{l}{\textbf{Accuracy}} & \multicolumn{1}{l}{\textbf{Entropy}} & \multicolumn{1}{l}{}                          \\ \midrule
\textcolor{black}{DR-CapsNet+Q ($\phi$)} & $90.51\pm_{0.29}$ & $6.25\pm_{1.15}$ & $77.40\pm_{0.64}$ & $5.01\pm_{1.45}$ & $0$ \\ 
\textcolor{black}{DR-CapsNet+REM ($\phi$)} & $90.38\pm_{0.53}$ & $3.35\pm_{1.18}$ & $76.98\pm_{0.45}$ & $2.47\pm_{0.96}$ & $50.13$ \\ \midrule
\textcolor{black}{DR-CapsNet+Q ($\psi$)} & $87.44\pm_{0.51}$ & $5.42\pm_{1.34}$ & $72.29\pm_{0.57}$ & $5.02\pm_{1.07}$ & $0$ \\
\textcolor{black}{DR-CapsNet+REM ($\psi$)} & $87.35\pm_{0.58}$ & $3.38\pm_{1.35}$ & $71.89\pm_{0.65}$ & $2.75\pm_{1.44}$ & $58.61$ \\\midrule
\textcolor{black}{$\gamma$-CapsNet+Q ($\phi$)} & $89.62\pm_{0.51}$ & $1.78\pm_{0.82}$ & $75.54\pm_{0.52}$ & $2.72\pm_{1.08}$ & $0$ \\
\textcolor{black}{$\gamma$-CapsNet+REM ($\phi$)} & $89.34\pm_{1.53}$ & $1.46\pm_{0.75}$ & $74.40\pm_{0.62}$ & $2.05\pm_{0.80}$ & $47.50$ \\\midrule
\textcolor{black}{$\gamma$-CapsNet+Q ($\psi$)} & $85.98\pm_{0.43}$ & $1.87\pm_{0.58}$ & $71.33\pm_{1.22}$ & $2.55\pm_{0.89}$ & $0$ \\ 
\textcolor{black}{$\gamma$-CapsNet+REM ($\psi$)} & $85.26\pm_{1.24}$ & $1.52\pm_{0.42}$ & $71.12\pm_{0.025}$ & $2.17\pm_{1.61}$ & $49.61$ \\ \midrule

\textcolor{black}{DeepCaps+Q ($\phi$)} & $95.01\pm_{0.58}$ & $7.32\pm_{1.18}$ & $83.18\pm_{1.61}$ & $7.28\pm_{1.66}$ & $0$ \\ 
\textcolor{black}{DeepCaps+REM ($\phi$)} & $94.62\pm_{0.52}$ & $6.75\pm_{1.41}$ & $82.49\pm_{1.53}$ & $6.12\pm_{1.91}$ & $34.45$ \\ \midrule
\textcolor{black}{DeepCaps+Q ($\psi$)} & $90.16\pm_{0.27}$ & $6.53\pm_{1.77}$ & $79.36\pm_{0.76}$ & $5.14\pm_{1.45}$  & $0$ \\
\textcolor{black}{DeepCaps+REM ($\psi$)} & $90.13\pm_{0.13}$ & $5.55\pm_{1.49}$ & $78.66\pm_{1.21}$ & $3.92\pm_{1.36}$ & $36.06$ \\ 
\bottomrule                      
\end{tabular}
\caption{\centering\label{tab:novelviewpoints_other}Accuracy (\%), entropy and sparsity (\%) on the smallNORB test set on familiar and novel viewpoints (elevations $\phi$ and azimuths $\psi$) for DR-CapsNets, $\gamma$-CapsNets and DeepCaps.}
\end{table}
We also show in Table~\ref{tab:novelviewpoints_full_original} the performances of these networks without quantization. All the models are trained with our implementations when the source code is unavailable.
\begin{table}[h]
\footnotesize
\centering
\begin{tabular}{lccc}
\toprule
\multicolumn{1}{c}{\textbf{Model}} & \textbf{Familiar} & \textbf{Novel} & \textbf{Sparsity} \\ \midrule
\textcolor{black}{DR-CapsNet ($\phi$)} & $90.62\pm_{0.21}$ & $77.51\pm_{0.43}$ & $0$ \\ 
\textcolor{black}{DR-CapsNet ($\phi$)} & $90.51\pm_{0.45}$ & $77.03\pm_{0.38}$ & $50.13$ \\ \midrule
\textcolor{black}{DR-CapsNet ($\psi$)} & $87.90\pm_{0.50}$ & $72.37\pm_{0.43}$ & $0$ \\
\textcolor{black}{DR-CapsNet ($\psi$)} & $86.81\pm_{0.61}$ &  $71.99\pm_{0.66}$ & $58.61$ \\\midrule
\textcolor{black}{$\gamma$-CapsNet ($\phi$)} & $90.15\pm_{0.39}$ & $75.89\pm_{0.53}$ & $0$ \\
\textcolor{black}{$\gamma$-CapsNet ($\phi$)} & $89.92\pm_{0.91}$ & $74.96\pm_{0.71}$  & $47.50$ \\\midrule
\textcolor{black}{$\gamma$-CapsNet ($\psi$)} & $86.11\pm_{0.63}$ & $72.55\pm_{0.83}$  & $0$ \\ 
\textcolor{black}{$\gamma$-CapsNet ($\psi$)} & $85.35\pm_{0.61}$ &  $71.35\pm_{1.31}$  & $49.61$ \\ \midrule

\textcolor{black}{DeepCaps ($\phi$)} & $95.32\pm_{0.48}$ &  $83.13\pm_{0.91}$ & $0$ \\ 
\textcolor{black}{DeepCaps ($\phi$)} & $94.48\pm_{0.31}$ &  $82.42\pm_{1.53}$ & $34.45$ \\ \midrule
\textcolor{black}{DeepCaps ($\psi$)} & $91.11\pm_{0.27}$ &  $79.53\pm_{0.78}$  & $0$ \\
\textcolor{black}{DeepCaps ($\psi$)} & $90.15\pm_{0.91}$ & $78.83\pm_{1.12}$ & $36.06$ \\ \midrule
\textcolor{black}{Eff-CapsNet ($\phi$)} & $97.83\pm_{0.41}$  & $85.04\pm_{1.06}$ & $0$ \\ 
\textcolor{black}{Eff-CapsNet ($\phi$)} & $98.10\pm_{0.33}$ &  $86.50\pm_{0.35}$ & $75.09$ \\ \midrule
\textcolor{black}{Eff-CapsNet ($\psi$)} & $97.07\pm_{0.52}$  & $87.98\pm_{1.64}$ & $0$ \\ 
\textcolor{black}{Eff-CapsNet ($\psi$)} & $97.29\pm_{0.34}$ &  $88.54\pm_{1.35}$ & $68.85$ \\ 
\bottomrule                      
\end{tabular}
\caption{\centering\label{tab:novelviewpoints_full_original}Accuracy (\%) and sparsity (\%) on the smallNORB test set on familiar and novel viewpoints (elevations $\phi$ and azimuths $\psi$) for DR-CapsNets, $\gamma$-CapsNets, DeepCaps and Eff-CapsNets without quantization.}
\end{table}

\subsection{Decoder and Limitations}\label{subsec:decoder}
\begin{figure}[h]
    \centering
    \includegraphics[width=0.6\columnwidth]{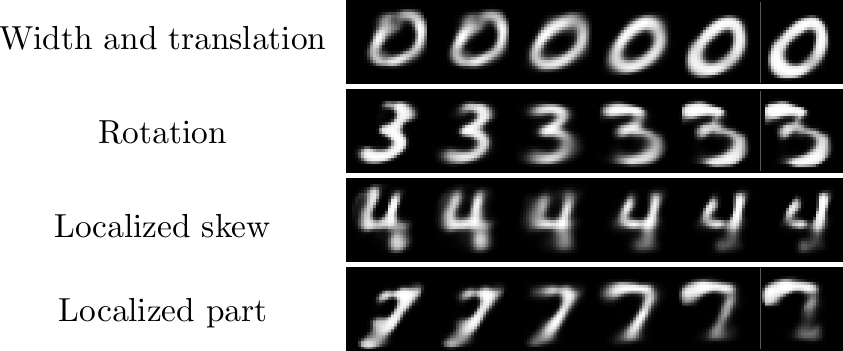}
    \caption{MNIST perturbation reconstructions of a frozen DR-CapsNet+REM.}
    \label{fig:reconstruction}
\end{figure}

A CapsNet is typically composed of an encoder and a decoder part, where the latter is a reconstruction network with three fully connected layers. In the previously discussed experiments, we have removed the decoder. One limitation of our work arises when computing the entropy of CapsNets trained with the decoder. 
Tables ~\ref{tab:decoder-acc-supp} and ~\ref{tab:decoder-ent-supp} report the classification results and entropies values when we trained the encoder and the decoder part together.
We observed that the entropy of a DR-CapsNets+REM is almost the same as that of a DR-CapsNet+Q. Indeed, when the decoder is used, the activity vector of an output capsule encodes richer representations of the input. The decoder was introduced to boost the routing performance on MNIST by enforcing the pose encoding a capsule. They also show that such perturbation affects the reconstruction when a perturbed activity vector is fed to the decoder. \textcolor{black}{So capsule representations are \emph{approximately equivariant}, meaning that even if they do not come with guaranteed equivariances, transformations applied to the input can still be described by continuous changes in the output vector.} To verify if output capsules of a trained DR-CapsNet+REM without the decoder (so with low entropy) are still approximately equivariant, we stacked on top of it the reconstruction network, without training the encoder. 
The MNIST dataset's decoder comprises three fully connected layers of 512, 1024, and 784 neurons, respectively, with two RELU and a final sigmoid activation function. This network is trained to minimize the Euclidean distance between the image and the output of the sigmoid layer.
We can see in Figure~\ref{fig:reconstruction} that DR-CapsNets+REM with low entropy are still approximately equivariant to many transformations. 
\begin{table}[h]
\footnotesize
\centering
\begin{tabular}{cccc}
\midrule
\textbf{Model}           & \textbf{MNIST}               & \textbf{F-MNIST}            & \textbf{CIFAR-10}            \\ \midrule
DR-CapsNet+Q      & $99.58\pm_{0.03}$ & $92.57\pm_{0.39}$ & $72.40\pm_{0.54}$ \\ 
DR-CapsNet+REM    & $99.63\pm_{0.02}$ & $92.76\pm_{0.38}$ & $76.00\pm_{0.63}$ \\ \midrule
\end{tabular}
\caption{\centering\label{tab:decoder-acc-supp}Classification results with the decoder on MNIST, Fashion-MNIST, CIFAR-10 (test set).}
\end{table}

\begin{table}[h]
\footnotesize
\centering
\begin{tabular}{cccc}
\midrule
\textbf{Model}        & \textbf{MNIST}             & \textbf{F-MNIST}           & \textbf{CIFAR-10}          \\ \midrule
DR-CapsNet+Q    & $9.88\pm_{0.06}$ & $8.49\pm_{1.50}$ & $4.55\pm_{1.13}$ \\
DR-CapsNets+REM & $9.40\pm_{0.55}$ & $6.15\pm_{2.32}$  & $3.85\pm_{0.54}$  \\ \midrule
\end{tabular}
\caption{\centering\label{tab:decoder-ent-supp}Entropies for models trained with the decoder on Fashion-MNIST and CIFAR-10 (test set).}
\end{table}

\subsection{Other distributions and tables}\label{sec:other_distr_tables}
\begin{figure}[h]
\centering
\includegraphics[width=0.6\columnwidth]{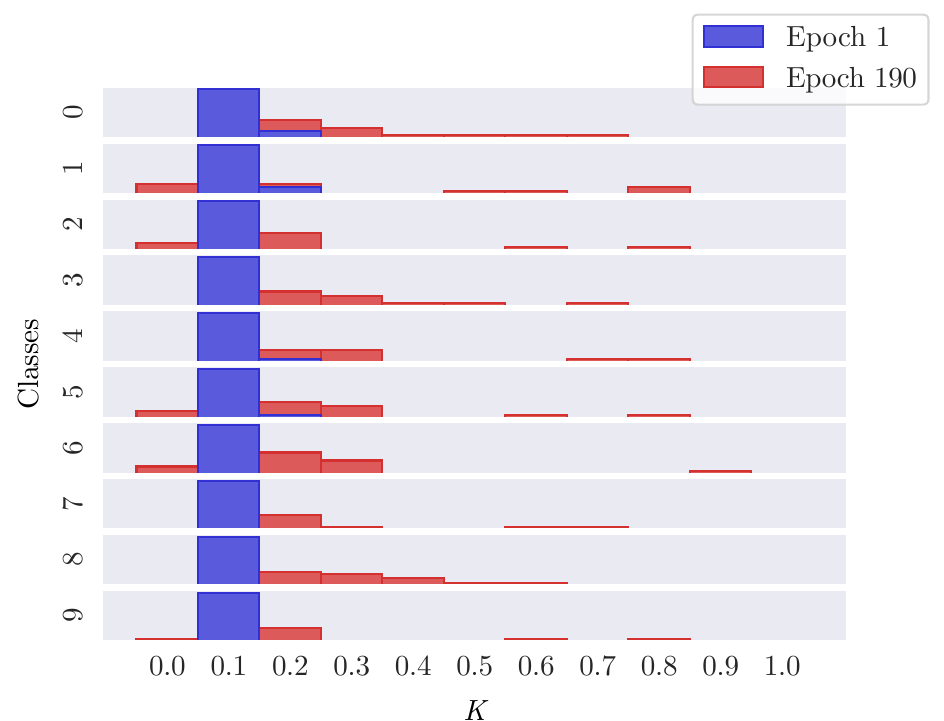}
\caption{Coupling coefficients distributions for each class of two DR-CapsNets+Q at epochs 1 and 190 on MNIST (test set).\label{fig:cij_nopruning_all}}
\end{figure}

Figure~\ref{fig:cij_nopruning_all} shows the distributions of the coupling coefficients for each class on MNIST of two DR-CapsNets+Q at epochs 1 and 190.

\begin{figure}[h]
      \centering
      \includegraphics[width=0.6\columnwidth]{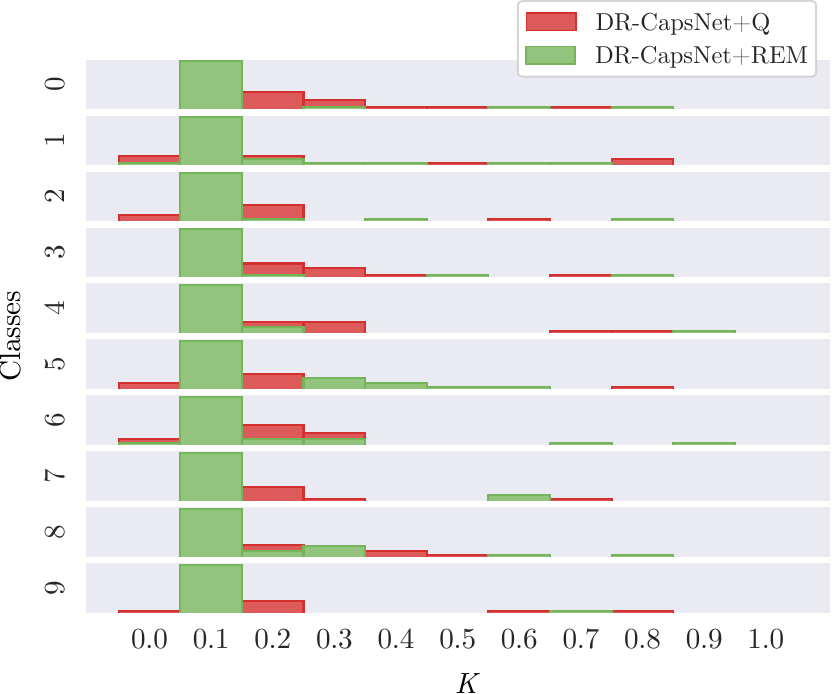}    \caption{\centering\label{fig:capsnetq_vscapsnetrem}Coupling coefficients distributions on MNIST (test set).}
\end{figure}

It can be observed that after the first epoch, DR-CapsNet is far from optimality, both in terms of performance (accuracy of 97.4\%) and parse tree discriminability: indeed, all coupling coefficients are almost equal to the value selected for initialization, i.e., $1/J$, where $J$ is the number of output capsules. 
Figure~\ref{fig:capsnetq_vscapsnetrem} illustrates the distributions of the coupling coefficients for a DR-CapsNet+Q and a DR-CapsNet+REM. We can see that the distributions of the DR-CapsNet+REM model are sparser than those for the DR-CapsNet+Q model. 
Table~\ref{tab:mnist-pruned-params} shows the performances on MNIST of DR-CapsNets, $\gamma$-CapsNets, DeepCaps and Eff-CapsNets. We notice that $\gamma$-CapsNet and $\gamma$-CapsNet+REM have the lowest entropy values since $\gamma$-CapsNets employ a scaled-distance-agreement routing algorithm that enforces the single parent constraint. With our REM technique, we can successfully lower the entropy even more. Table~\ref{tab:mnist-pruned-params} includes the corresponding number of trainable parameters for completeness.
\begin{table}[h]
    \footnotesize
    \centering
    \begin{tabular}{l*{4}{c}}
        \toprule
        \textbf{Model}  &\textbf{Accuracy}   & \textbf{Entropy} & \textbf{Sparsity} & \textbf{Parameters}\\ \midrule
        DR-CapsNet           & $99.57\pm_{0.02}$ & $-$              & $-$     & 6.8M \\
        DR-CapsNet+Q         & $99.56\pm_{0.03}$ & $9.53\pm_{0.54}$ & $-$     & 6.8M \\
        DR-CapsNet+REM       & $99.56\pm_{0.02}$ & $4.16\pm_{1.59}$ & $85.53$ & 0.9M \\
        $\gamma$-CapsNet     & $99.53\pm_{0.15}$ & $-$              & $-$     & 7.7M \\
        $\gamma$-CapsNet+Q   & $99.50\pm_{0.07}$ & $1.87\pm_{1.38}$ & $-$     & 7.7M \\
        $\gamma$-CapsNet+REM & $99.50\pm_{0.05}$ & $1.34\pm_{1.09}$ & $89.71$ & 0.8M \\
        DeepCaps             & $99.58\pm_{0.32}$ & $-$              & $-$     & 8.4M \\
        DeepCaps+Q           & $99.51\pm_{0.24}$ & $5.26\pm_{2.00}$ & $-$     & 8.4M \\
        DeepCaps+REM         & $99.61\pm_{0.23}$ & $3.10\pm_{1.07}$ & $71.73$ & 2.4M \\
        Eff-CapsNet          & $99.57\pm_{0.48}$ & $-$              & $-$     & 161k \\
        Eff-CapsNet+Q        & $99.55\pm_{0.31}$ & $4.38\pm_{1.59}$ & $-$     & 161k \\
        Eff-CapsNet+REM      & $99.58\pm_{0.64}$ & $2.60\pm_{1.72}$ & $73.15$ & 43k  \\\bottomrule
    \end{tabular}
    \caption{\centering\label{tab:mnist-pruned-params}Accuracy (\%), entropy, sparsity (\%) and corresponding number of trainable parameters for DR-CapsNets, $\gamma$-CapsNets, DeepCaps and Eff-CapsNets on MNIST (test set).}
\end{table}
Table~\ref{tab:results-ti} reports the accuracy, sparsity, and entropy values for DR-CapsNets on Tiny ImageNet.
\begin{table}[h]
    \footnotesize
    \centering
    \footnotesize
    \begin{tabular}{lccc}
        \toprule
        \textbf{Model} & \textbf{Accuracy} & \textbf{Entropy} & \textbf{Sparsity}  \\ \midrule
        DR-CapsNet     & $60.85\pm_{1.91}$ & -                & $0$      \\
        DR-CapsNet+Q   & $58.50\pm_{1.46}$ & $5.18\pm_{0.67}$ & $0$     \\
        DR-CapsNet+REM & $58.34\pm_{1.61}$ & $3.15\pm_{0.81}$ & $44.27$ \\ \midrule
    \end{tabular}\caption{\centering\label{tab:results-ti}Accuracy (\%), entropy and sparsity (\%) results on Tiny ImageNet (test set).}
\end{table}

Table~\ref{tab:num_parse_trees_all} reports the number of parse trees for Fashion-MNIST, SVHN, CIFAR-10, affNIST, and Tiny ImageNet (only the first ten classes).
\begin{table}[h]
\footnotesize
\centering
\begin{tabular}{*{13}{c}}
\toprule
\textbf{Class} & \multicolumn{2}{c}{F-MNIST} & \multicolumn{2}{c}{affNIST} & \multicolumn{2}{c}{CIFAR-10} & \multicolumn{2}{c}{T-ImageNet} & \multicolumn{2}{c}{SVHN} & \\
 & Q & REM & Q & REM &Q & REM &Q & REM & Q & REM\\
 \midrule
$\mathbf{\textbf{\#}_0}$ & $610$ & $58$  & $9640$  & $248$& $345$ & $64$ & $65$ & $28$  & \textcolor{black}{$585$} & \textcolor{black}{$74$}\\
$\mathbf{\textbf{\#}_1}$ &$936$ & $140$ & $5490$ & $66$ & $407$& $125$ & $57$& $22$ & \textcolor{black}{$1750$} & \textcolor{black}{$85$}\\
$\mathbf{\textbf{\#}_2}$ &$332$ & $35$ & $14055$ &$438$ & $375$ & $115$ & $50$ & $9$ & \textcolor{black}{$1985$} & \textcolor{black}{$302$}\\
$\mathbf{\textbf{\#}_3}$ &$600$ & $80$ & $4446$ & $97$ & $200$ & $70$ & $12$ & $11$ & \textcolor{black}{$1054$} & \textcolor{black}{$266$}\\
$\mathbf{\textbf{\#}_4}$ &$346$ & $30$ & $9059$ & $161$ & $291$ & $39$ & $39$ & $15$ & \textcolor{black}{$890$} & \textcolor{black}{$60$}\\
$\mathbf{\textbf{\#}_5}$ &$828$ & $35$ & $7732$ &$425$ & $235$ & $76$ & $51$ & $7$ & \textcolor{black}{$1116$} & \textcolor{black}{$129$ }\\
$\mathbf{\textbf{\#}_6}$ &$297$ & $23$ & $11957$ & $1244$ &$282$ &$58$ & $90$ & $14$  & \textcolor{black}{$780$} & \textcolor{black}{$70$}\\
$\mathbf{\textbf{\#}_7}$ &$915$ & $31$ & $3109$ & $164$ & $381$ & $71$ & $48$ & $21$ & \textcolor{black}{$676$} & \textcolor{black}{$31$}\\
$\mathbf{\textbf{\#}_8}$ &$812$ & $217$ & $15521$ & $396$ &$305$ &$37$ & $129$ &$30$ & \textcolor{black}{$434$} & \textcolor{black}{$139$}\\
$\mathbf{\textbf{\#}_9}$ &$978$&$615$&$3703$& $67$ &$147$&$52$ & $35$& $12$  & \textcolor{black}{$681$} & \textcolor{black}{$110$}\\ 
\midrule
\end{tabular}
\caption{\centering\label{tab:num_parse_trees_all}Number of parse trees for each class of a DR-CapsNet+Q and DR-CapsNet+REM on Fashion-MNIST, affNIST, CIFAR-10, Tiny ImageNet (first ten classes) and SVHN.}
\end{table}

{\color{ForestGreen}As mentioned in the ablation study in Section~\ref{sec:abl}, we can plug any pruning strategy with REM. However, due to better performance in terms of accuracy and entropy values, iterative magnitude pruning~\citep{lobster} resulted in a suitable choice to enforce REM. In fact, in Figure~\ref{fig:snip-cifar10}, we can see that employing a single-shot pruning strategy such as SNIP~\citep{lee2018snip} leads to higher entropy and lower accuracy even for more complex datasets than MNIST, such as CIFAR-10, compared to the results presented in Table~\ref{tab:results-otherdatasets}.
 \begin{figure}[h!]
 \centering
 \includegraphics[width=0.8\columnwidth]{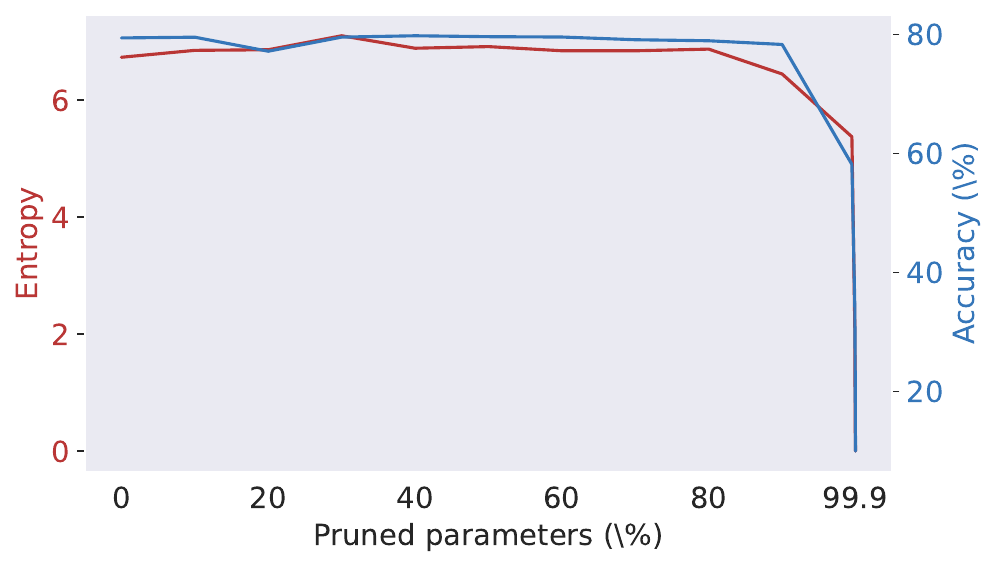}
 \caption{\centering\label{fig:snip-cifar10}\textcolor{ForestGreen}{Entropy (red line) and accuracy (blue line) of DR-CapsNe+REM at different epochs (with single-shot pruning) on CIFAR-10 (test set).}}
 \end{figure}

See also Table~\ref{tab:snip-cifar10} for a better comparison with similar sparsity.
\begin{table}[h]
    \footnotesize
    \centering
    \footnotesize
    \begin{tabular}{lcccc}
        \toprule
        \textbf{Model} & \textbf{Pruning Strategy} & \textbf{Accuracy} & \textbf{Entropy} & \textbf{Sparsity}  \\ \midrule
        DR-CapsNet+REM & iterative & $78.95\pm_{0.08}$ & $6.87\pm_{1.08}$ & $80.00$     \\
        DR-CapsNet+REM & single-shot & $79.25\pm_{0.59}$ & $4.15\pm_{0.62}$ & $81.17$ \\ \midrule
    \end{tabular}\caption{\centering\label{tab:snip-cifar10}\textcolor{ForestGreen}{Accuracy (\%), entropy and sparsity (\%) results on CIFAR-10 (test set) when using REM with iterative and single-shot pruning.}}
\end{table}}

\subsection{Additional visualizations}\label{sec:other_visualizations}
\textcolor{ForestGreen2}{The threshold mentioned in Section~\ref{sec:extract_salmap} is used exclusively for visualization purposes, helping to highlight the most significant connections in the parse trees of fully connected capsules. It does not affect the performance of our proposed algorithm. Specifically, we set the threshold to $1/|\Omega^{[l]}|$, where $|\Omega^{[l]}|$ represent the number of capsules in layer $l$. This choice provides a good trade-off between clarity and readability in visualizations. Importantly, this threshold is not static — it naturally adapts when using different network architectures since it varies accordingly. Since the threshold is only used for visualization, there is no need for extensive tuning when applying our method to new backbones. However, if one wishes to adjust it for specific visualization preferences, it can be empirically chosen based on the level of detail desired in the parse tree representation.}
\subsubsection{Fashion-MNIST parse trees}
\begin{figure}[h]
\centering
         \includegraphics[width=0.7\columnwidth]{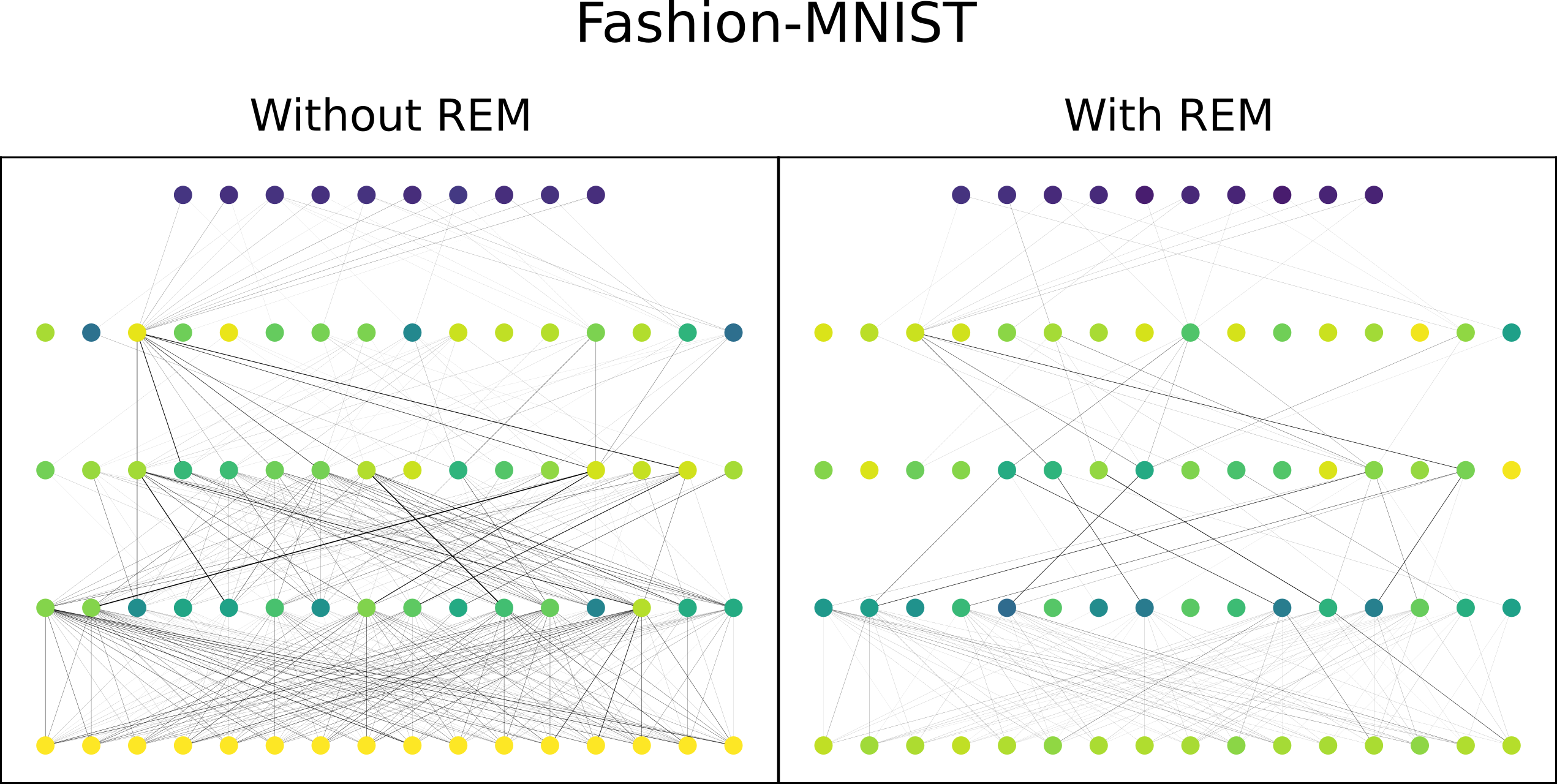}
\caption{\textcolor{black}{Parse trees for the entire Fashion-MNIST test set for multilayered DR-CapsNet and DR-CapsNet+REM with backtracking.}}
\label{fig:parse_trees_fm_dataset}
\end{figure}

\begin{figure}[h]
\centering
         \includegraphics[width=1\columnwidth]{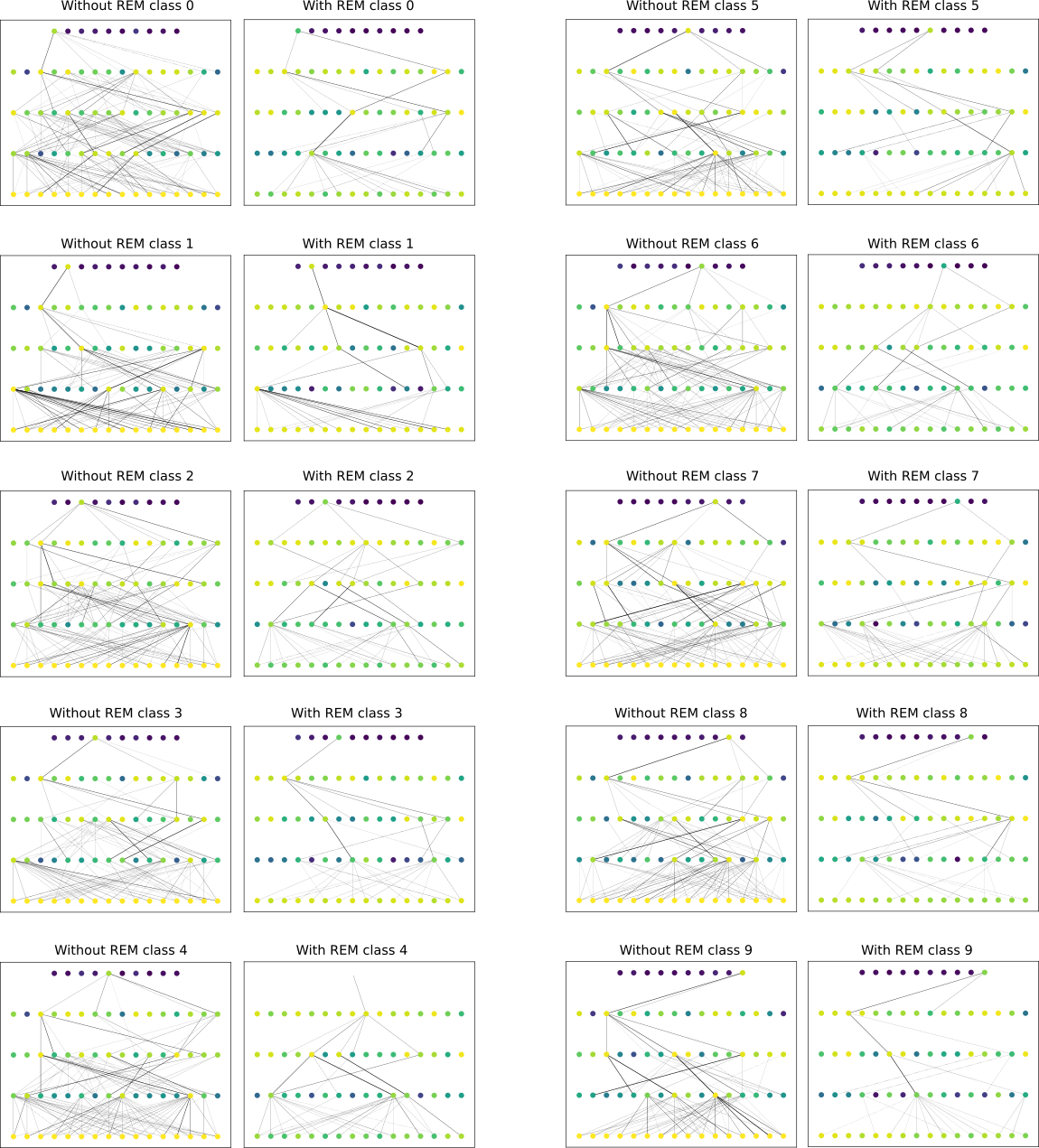}
\caption{\textcolor{black}{Parse trees for each class of Fashion-MNIST (test set) for multilayered DR-CapsNet and DR-CapsNet+REM with backtracking.}}
\label{fig:parse_trees_fm_classes}
\end{figure}

\begin{figure}[h]
\centering
         \includegraphics[width=1\columnwidth]{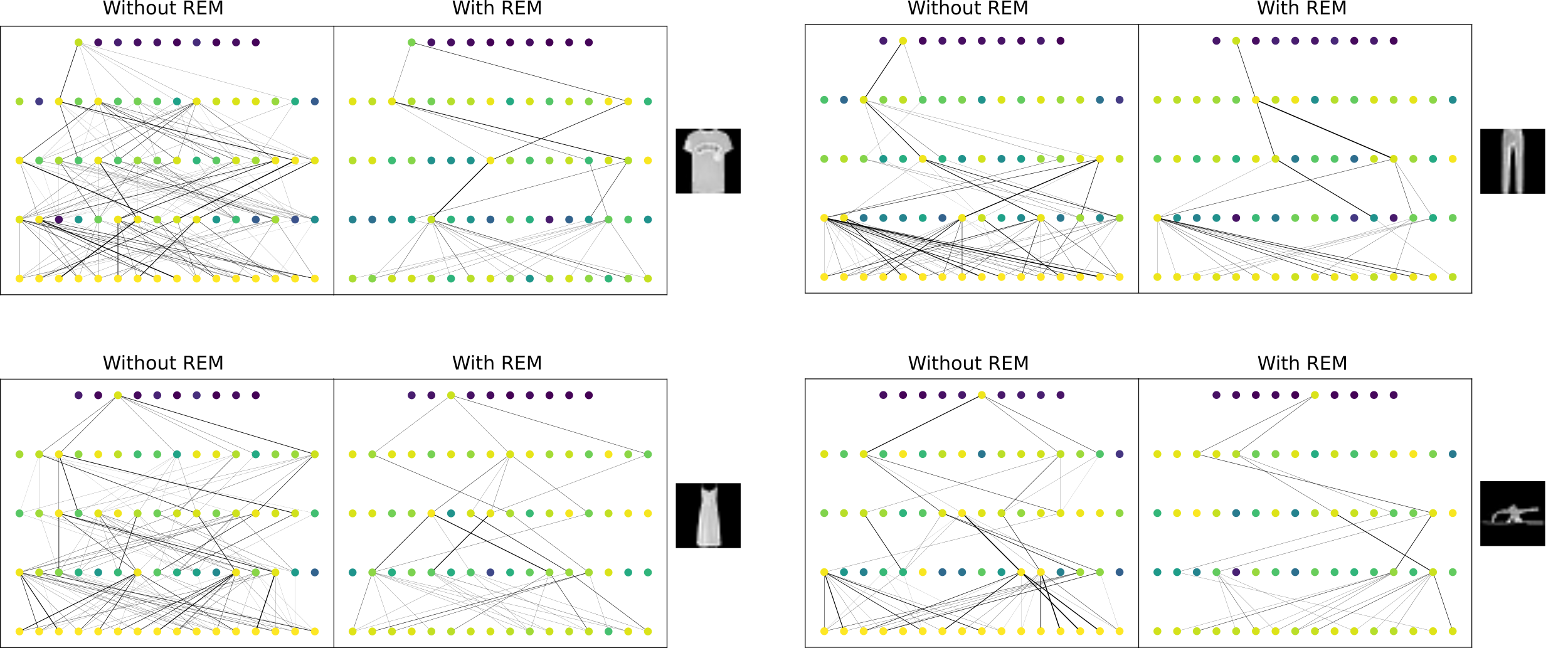}
\caption{\textcolor{black}{Parse trees for some Fashion-MNIST test images for multilayered DR-CapsNet and DR-CapsNet+REM with backtracking.}}
\label{fig:parse_trees_fm_samples}
\end{figure}

In Figure~\ref{fig:parse_trees_fm_dataset}, we show additional parse trees of multilayered DR-CapsNets for all the images of the test set of Fashion-MNIST (obtained by averaging the activations and coupling coefficients of all images), in Figure~\ref{fig:parse_trees_fm_classes} the parse trees for each class and in Figure~\ref{fig:parse_trees_fm_samples} the parse trees for some test samples.
\subsubsection{CIFAR-10 parse trees}
\begin{figure}[h]
\centering
         \includegraphics[width=0.7\columnwidth]{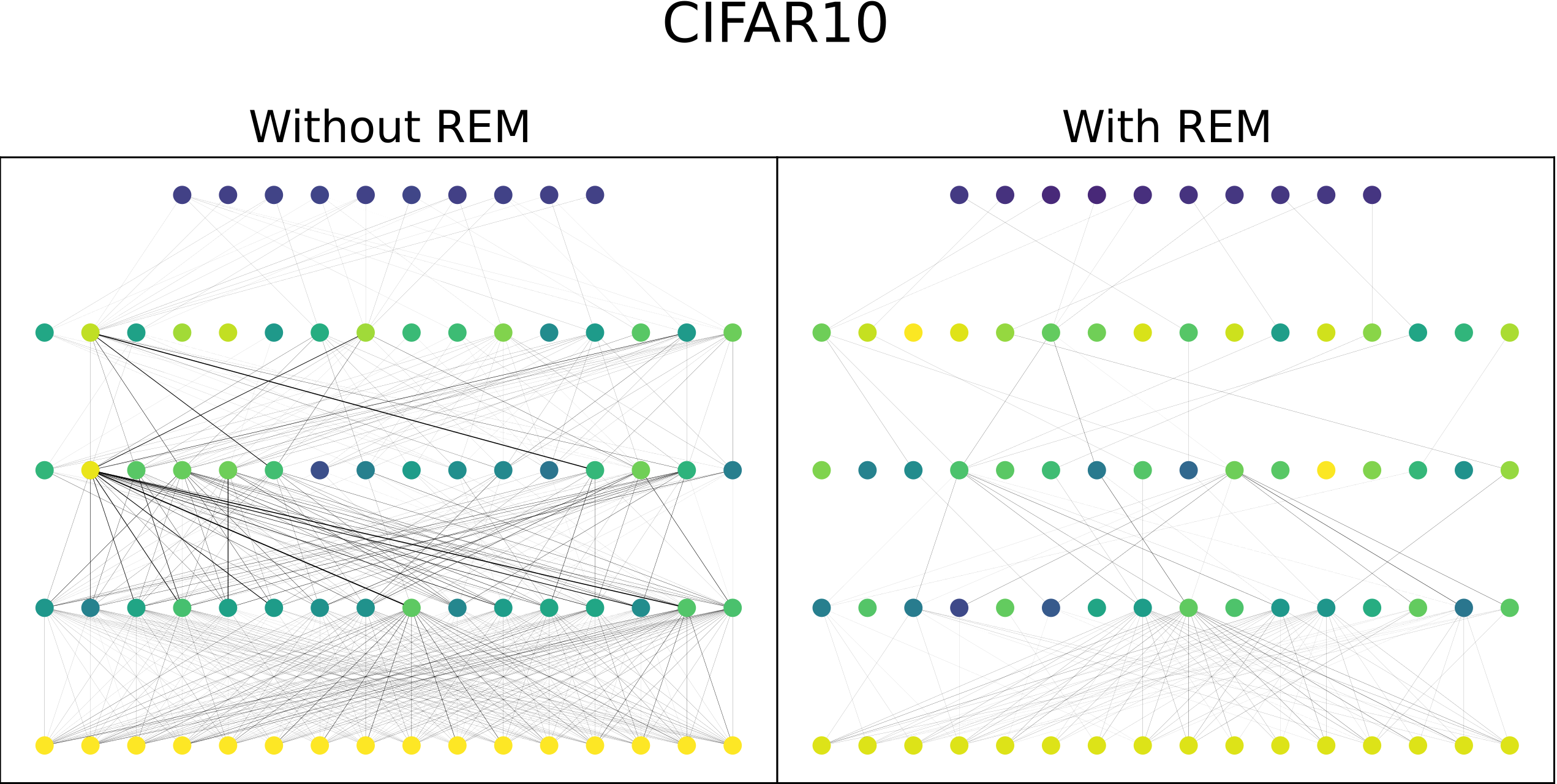}
\caption{\textcolor{black}{Parse trees for the entire CIFAR-10 test set for multilayered DR-CapsNet and DR-CapsNet+REM with backtracking.}}
\label{fig:parse_trees_cifar10_dataset}
\end{figure}

\begin{figure}[h]
\centering
         \includegraphics[width=1\columnwidth]{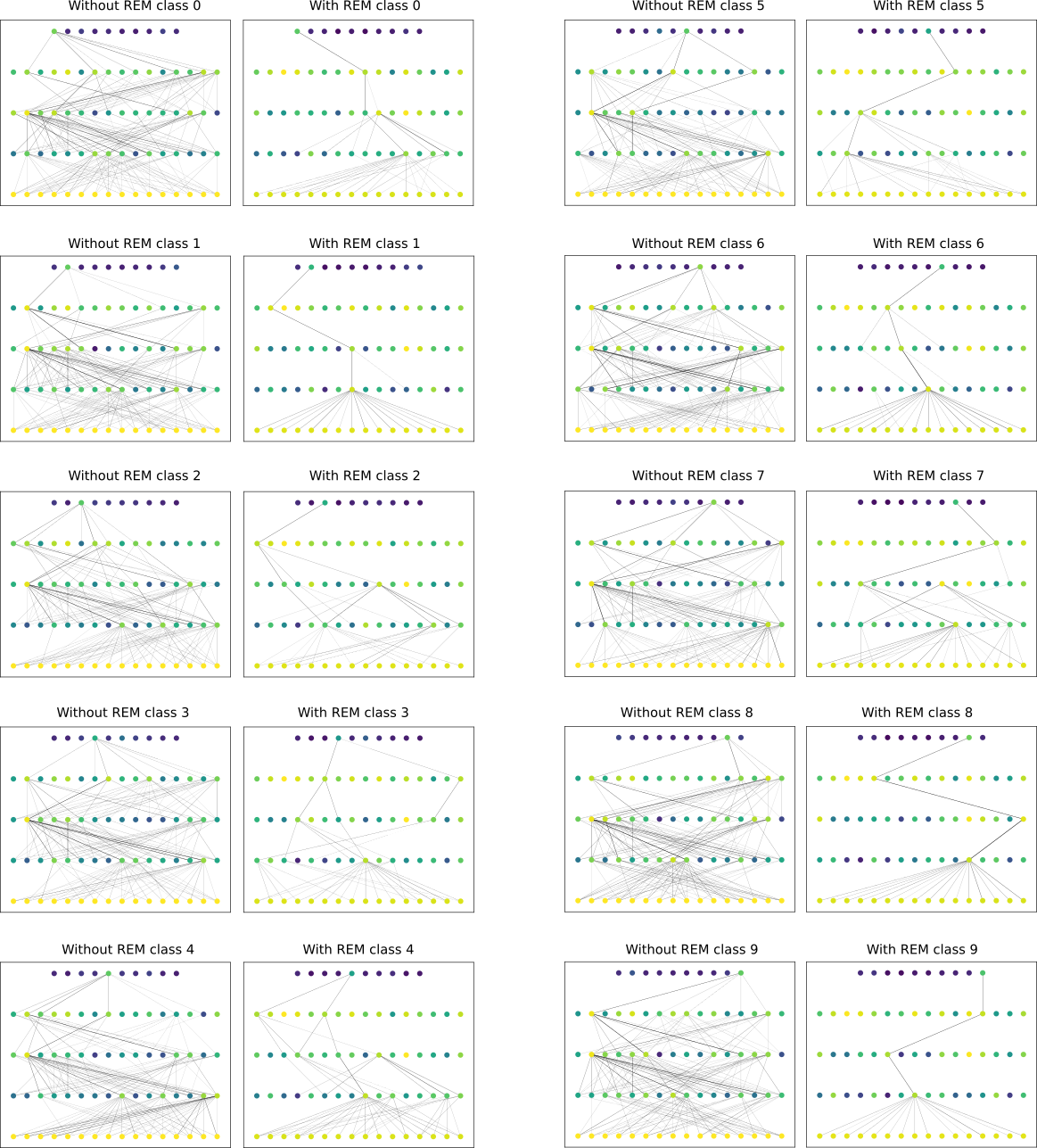}
\caption{\textcolor{black}{Parse trees for each class of CIFAR-10 (test set) for multilayered DR-CapsNet and DR-CapsNet+REM with backtracking.}}
\label{fig:parse_trees_cifar10_classes}
\end{figure}

\begin{figure}[h]
\centering
         \includegraphics[width=0.9\columnwidth]{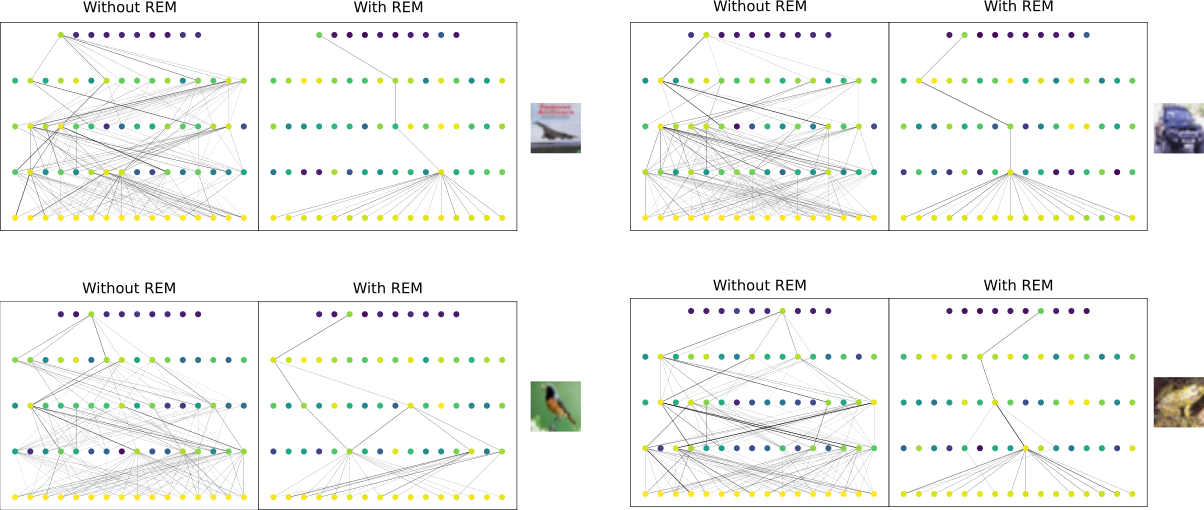}
\caption{\textcolor{black}{Parse trees for some CIFAR-10 test images for multilayered DR-CapsNet and DR-CapsNet+REM with backtracking.}}
\label{fig:parse_trees_cifar10_samples}
\end{figure}

In Figure~\ref{fig:parse_trees_cifar10_dataset}, we show additional parse trees of multilayered DR-CapsNets for all the images of the test set of CIFAR-10 (obtained by averaging the activations and coupling coefficients of all images), in Figure~\ref{fig:parse_trees_cifar10_classes} the parse trees for each class and in Figure~\ref{fig:parse_trees_cifar10_samples} the parse trees for some test samples.

\subsubsection{smallNORB saliency maps}
\begin{figure}[h]
\centering
         \includegraphics[width=0.7\columnwidth]{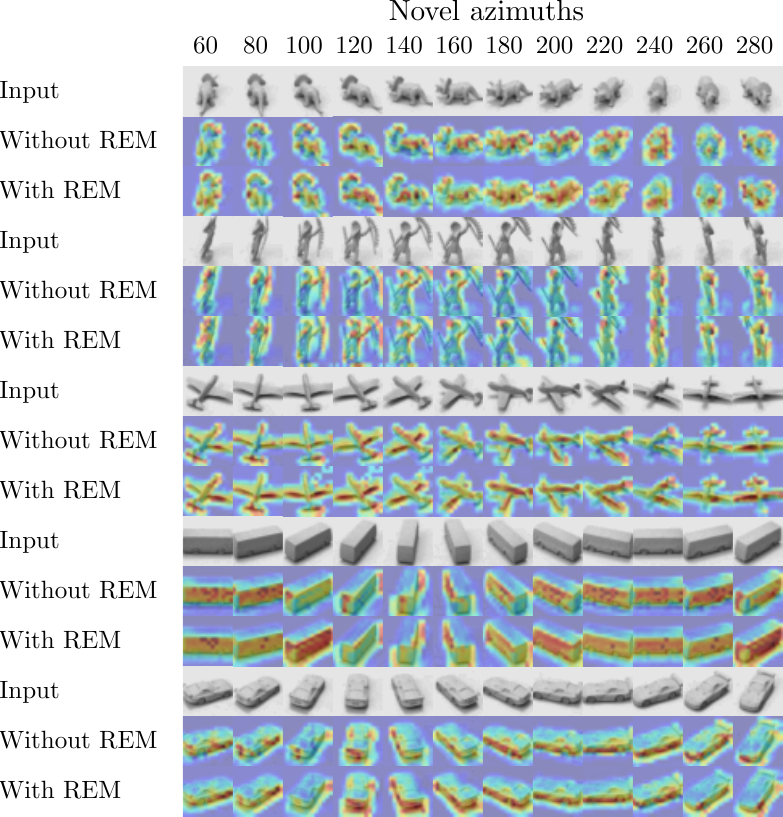}
\caption{\textcolor{black}{Saliency maps for smallNORB for DR-CapsNet+Q and DR-CapsNet+REM (novel azimuths).}}
\label{fig:saliency_maps_sn_az}
\end{figure}

\begin{figure}[h]
\centering
         \includegraphics[width=0.45\columnwidth]{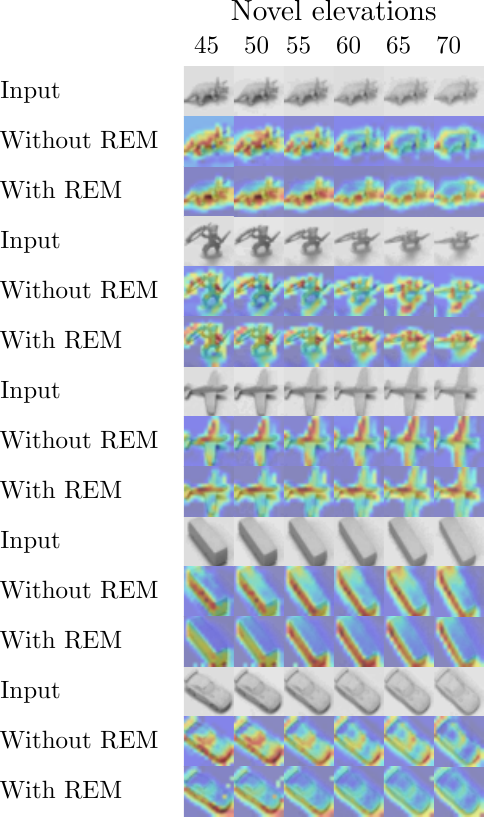}
\caption{\textcolor{black}{Saliency maps for smallNORB for DR-CapsNet+Q and DR-CapsNet+REM (novel elevations).}}
\label{fig:saliency_maps_sn_el}
\end{figure}

Here we show in Figures~\ref{fig:saliency_maps_sn_az} and \ref{fig:saliency_maps_sn_el} the saliency maps for smallNORB when tested on novel azimuths and elevations, respectively. We used the method described in ~\ref{sec:extract_salmap}. We show one object instance for each category. We can see from these visualizations on smallNORB that the differences between saliency maps of REM and without REM are less prominent. This is because smallNORB is less complex than Fashion-MNIST or CIFAR-10. However, we can still notice that some of the saliency maps extracted with REM are more prominent and consistent between each category.

\clearpage
\newpage
\textcolor{ForestGreen2}{\section{Derivation of $\frac{\partial \mathcal{L}}{\partial \mathbf{W}_{i,j}}$}\label{app:derivation}}

\noindent The votes vectors $\hat{\mathbf{u}}_{i,j}$ of capsules $i$ in layer $l$ for capsules $j$ in layer $l+1$ are computed as:
\begin{equation}
    \hat{\mathbf{u}}_{i,j} = \mathbf{W}_{i,j} \mathbf{u}_i^{^{(r_T)}},
\end{equation}
where $\mathbf{u}_i^{^{(r_T)}}$ is the pose of capsule $i$ at the last routing iteration $r_T$.

\noindent The routing process involves computing the unnormalized capsule pose
\begin{equation}
    \mathbf{s}_j^{^{(r_T)}} = \sum_i c_{i,j}^{^{(r_T)}} \hat{\mathbf{u}}_{i,j},
\end{equation}
where $c_{i,j}^{(r_T)}$ are the routing coefficients.

\noindent The squashing function is applied to obtain the final capsule activation:
\begin{equation}
    \mathbf{u}_j^{^{(r_T)}} = \squash(\mathbf{s}_j^{^{(r_T)}}) = \frac{\|\mathbf{s}_j^{^{(r_T)}}\|^2}{1 + \|\mathbf{s}_j^{^{(r_T)}}\|^2} \frac{\mathbf{s}_j^{^{(r_T)}}}{\|\mathbf{s}_j^{^{(r_T)}}\|}.
\end{equation}

\noindent The routing coefficients $c_{i,j}^{^{(r_T)}}$ are computed via a softmax function:
\begin{equation}
    c_{i,j}^{^{(r_T)}} = \frac{\exp(b_{i,j}^{^{(r_T-1)}})}{\sum_k \exp(b_{ik}^{^{(r_T-1)}})},
\end{equation}
where the logits $b_{i,j}^{^{(t)}}$ are updated iteratively as:
\begin{equation}
    b_{i,j}^{^{(t)}} = b_{i,j} + \sum_{r=1}^{t-1} \mathbf{u}_j^{(r)} \cdot \hat{\mathbf{u}}_{i,j},
\end{equation}
and $b_{i,j}$ are the initial logits before routing.

\noindent The loss function $\mathcal{L}$ depends on the final output capsules $\mathbf{u}_j^{(r_T)}$ after $r_T$ iterations.

\noindent Using the chain rule and following~\citet{gu}, we expand:
\begin{equation}
\label{eq:loss_votes}
    \frac{\partial \mathcal{L}}{\partial \hat{\mathbf{u}}_{i,j}} =
    \frac{\partial \mathcal{L}}{\partial \mathbf{u}_j^{(r_T)}} 
    \frac{\partial \mathbf{u}_j^{(r_T)}}{\partial \mathbf{s}_j^{(r_T)}}
    \frac{\partial \mathbf{s}_j^{(r_T)}}{\partial \hat{\mathbf{u}}_{i,j}}
    +
    \sum_{m=1}^{J} 
    \frac{\partial \mathcal{L}}{\partial \mathbf{u}_m^{(r_T)}}
    \frac{\partial \mathbf{u}_m^{(r_T)}}{\partial \mathbf{s}_m^{(r_T)}}
    \frac{\partial \mathbf{s}_m^{(r_T)}}{\partial c_{i,m}^{(r_T)}}
    \frac{\partial c_{i,m}^{(r_T)}}{\partial \hat{\mathbf{u}}_{i,j}}.
\end{equation}

\noindent There are \textbf{two terms} in the gradient because:
\begin{enumerate}
    \item $\hat{\mathbf{u}}_{i,j}$ \textbf{directly affects} $\mathbf{s}_j$, and hence $\mathbf{u}_j$.
    
    \item $\hat{\mathbf{u}}_{i,j}$ \textbf{indirectly affects} all routing logits $b_{i,m}$, and hence all routing coefficients $c_{i,m}$, and hence all $\mathbf{s}_m$, which affect all $\mathbf{u}_m$.
\end{enumerate}

\noindent  In fact, we can see that
\begin{equation}
    c_{i,j}^{^{(r_T)}} = \frac{\exp\left(b_{i,j} + \sum_{r=1}^{r_T-2} \boldsymbol{u}_j^{(r)} \hat{\mathbf{u}}_{i,j}\right)}{\sum_m^J \exp\left(b_{i,m} + \sum_{r=1}^{r_T-2}\boldsymbol{u}_m^{(r)}\hat{\mathbf{u}}_{i,m}\right)},
\end{equation}

\noindent So we need \textbf{both paths} when computing the total gradient through routing. Let's go through the derivation of $\frac{\partial \mathcal{L}}{\partial \mathbf{W}_{i,j}}$.

\noindent\textbf{First term of the sum.} Differentiating $\mathbf{s}_j$ w.r.t. $\hat{\mathbf{u}}_{i,j}$ gives:
\begin{equation}
    \frac{\partial \mathbf{s}_j^{^{(r_T)}}}{\partial \hat{\mathbf{u}}_{i,j}} = c_{i,j}^{^{(r_T)}},
\end{equation}

\noindent So we obtain:
\begin{equation}
    \frac{\partial \mathcal{L}}{\partial \mathbf{u}_j^{(r_T)}}
    \frac{\partial \mathbf{u}_j^{(r_T)}}{\partial \mathbf{s}_j^{(r_T)}}
    c_{i,j}^{^{(r_T)}},
\end{equation}
which is the expanded first term of the sum of Equation~\ref{eq:loss_votes}.

\noindent\noindent\textbf{Second term of the sum.} Differentiating $\mathbf{s}_m$ w.r.t. $c_{i,m}$ gives: 
\begin{equation}
    \frac{\partial \mathbf{s}_m^{^{(r_T)}}}{\partial c_{i,m}^{^{(r_T)}}} = \hat{\mathbf{u}}_{i,m}.
\end{equation}

\noindent Thus,
\begin{equation}
    \sum_{m=1}^{J} 
    \frac{\partial \mathcal{L}}{\partial \mathbf{u}_m^{(r_T)}}
    \frac{\partial \mathbf{u}_m^{(r_T)}}{\partial \mathbf{s}_m^{(r_T)}}
    \frac{\partial \mathbf{s}_m^{(r_T)}}{\partial c_{i,m}^{(r_T)}}
    \frac{\partial c_{i,m}^{(r_T)}}{\partial \hat{\mathbf{u}}_{i,j}} = 
    \sum_{m=1}^{J} 
    \frac{\partial \mathcal{L}}{\partial \mathbf{u}_m^{(r_T)}}
    \frac{\partial \mathbf{u}_m^{(r_T)}}{\partial \mathbf{s}_m^{(r_T)}}
    \hat{\mathbf{u}}_{i,m} \frac{\partial c_{i,m}^{(r_T)}}{\partial \hat{\mathbf{u}}_{i,j}}
\end{equation}
yields the second term in the final equation.

\noindent The final gradient of the loss w.r.t. $\hat{\mathbf{u}}_{i,j}$ is
\begin{equation}
\begin{aligned}
    \frac{\partial \mathcal{L}}{\partial \hat{\mathbf{u}}_{i,j}} 
    & = \frac{\partial \mathcal{L}}{\partial \mathbf{u}_j^{(r_T)}} 
    \frac{\partial \mathbf{u}_j^{(r_T)}}{\partial \mathbf{s}_j^{(r_T)}}
    c_{i,j}^{^{(r_T)}}
    +
    \sum_{m=1}^{J} 
    \frac{\partial \mathcal{L}}{\partial \mathbf{u}_m^{(r_T)}}
    \frac{\partial \mathbf{u}_m^{(r_T)}}{\partial \mathbf{s}_m^{(r_T)}}
    \hat{\mathbf{u}}_{i,m}\frac{\partial c_{i,m}^{(r_T)}}{\partial \hat{\mathbf{u}}_{i,j}}.
\end{aligned}
\end{equation}

\noindent Using:
\begin{equation}
    \frac{\partial \hat{\mathbf{u}}_{i,j}}{\partial \mathbf{W}_{i,j}} = \mathbf{u}_i,
\end{equation}
we obtain:
\begin{equation}
    \frac{\partial \mathcal{L}}{\partial \mathbf{W}_{i,j}} =
    \frac{\partial \mathcal{L}}{\partial \hat{\mathbf{u}}_{i,j}} \mathbf{u}_i.
\end{equation}

\end{document}